\documentclass[twoside,11pt]{article}
\usepackage{jmlr2e}
\usepackage{amsmath}

\usepackage[T1]{fontenc}
\usepackage[latin9]{inputenc}
\usepackage{comment}
\usepackage{bm}
\usepackage{amstext}

\usepackage{amssymb}
\newcommand{\qed}{\hfill $\square$}
\usepackage{amsfonts}
\usepackage{bbm}

\usepackage{graphicx}
\usepackage{color}

\newtheorem{assumption}{Assumption}

\renewcommand{\hat}{\widehat}

\newcommand{\E}{\mathbb{E}}

\newcommand{\Var}{\mathbb{V}}

\usepackage{algorithm}
\usepackage{algorithmic}
\begin{document}

\ShortHeadings{Empirical Likelihood for Random Forests}{Chiang, Matsushita and Otsu}
\firstpageno{1}

\title{Empirical Likelihood for Random Forests \\
and Ensembles}

\author{\name Harold D. Chiang\\
  \addr Department of Economics \\
        University of Wisconsin--Madison \\
        1180 Observatory Drive \\
        Madison, WI 53706--1393, USA
  \email hdchiang@wisc.edu
  \AND
  \name Yukitoshi Matsushita\\
  \addr Graduate School of Economics \\
        Hitotsubashi University \\
        2-1 Naka, Kunitachi \\
        Tokyo 186-8601, Japan
  \email matsushita.y@r.hit-u.ac.jp
  \AND
  \name Taisuke Otsu\\
  \addr Department of Economics \\
        London School of Economics \\
        Houghton Street \\
        London WC2A 2AE, UK
  \email t.otsu@lse.ac.uk
}

\date{}
\editor{}
\maketitle

\begin{abstract}
We develop an empirical likelihood (EL) framework for random forests and related ensemble methods, providing a likelihood-based approach to quantify their statistical uncertainty. Exploiting the incomplete $U$-statistic structure inherent in ensemble predictions, we construct an EL statistic that is asymptotically chi-squared when subsampling induced by incompleteness is not overly sparse. Under sparser subsampling regimes, the EL statistic tends to over-cover due to loss of pivotality; we therefore propose a modified EL that restores pivotality through a simple adjustment. Our method retains key properties of EL while remaining computationally efficient. Theory for honest random forests and simulations demonstrate that modified EL achieves accurate coverage and practical reliability relative to existing inference methods.
\end{abstract}


\allowdisplaybreaks

\section{Introduction}
A random forest \citep{breiman2001random} is a powerful and versatile machine-learning method. In contrast with many other predictive algorithms, it requires minimal tuning yet yields strong performance even in moderate samples. Its capacity to handle high-dimensional feature spaces also renders them considerably more flexible than conventional nonparametric estimators. A comprehensive survey is provided by \cite{biau2016random}.

At a high level, a random forest operates via  a ``divide-and-conquer" principle: it is an ensemble of regression or classification trees, each trained on a resampled version of the data, with or without replacement. The principle of aggregating predictions from multiple base learners trained on subsamples is general and has inspired a wide class of ensemble methods. While these procedures often excel in prediction, quantifying the associated statistical uncertainty remains challenging. In contrast to conventional statistical methods,  well-founded likelihood-based inferential tools are largely unavailable for ensembles like random forests. This paper develops an empirical likelihood (EL) framework \citep{Owen1988} for random forests and related ensemble learners, thereby enabling likelihood-based inference for this widely used class of models.

Despite recent progress in the asymptotic theory and inference on random forests, many existing analyses either neglect the resampling stage and the additional uncertainty it induces, or only partially account for its influence on the asymptotic distribution of predictions, while leaving its impact on variance estimation unaddressed. Moreover, to the best of our knowledge, no EL or other related likelihood-based inference method has yet been developed in this context.

Random forests and related ensemble methods possess an underlying $U$-statistic structure, suggesting that the jackknife EL approach of \cite*{jing2009jackknife} could, in principle, be applied directly. However, because their asymptotic theory is developed only for finite-order $U$-statistics, it is unclear whether these results extend to the infinite-order setting induced by random forests, where often the order of the kernel must grow rapidly with the sample size for the random forest  prediction bias  to vanish asymptotically--see e.g., Theorem 3 in \cite{wager2018estimation}. More fundamentally, complete random forests and their leave-one-out versions  are computationally prohibitive even for moderate sample sizes. Without them, the jackknife pseudo-values cannot be formed, rendering the jackknife EL procedure inapplicable in practice.

In practice, random forest predictions correspond to an incomplete $U$-statistic: the ensemble is formed from only a small subset of all possible subsample-based trees, rather than the full combinatorial collection. To address this computational bottleneck, we propose a new EL inference method that builds on the principle of jackknife EL and incorporates a jackknife-after-subsampling scheme--an analogue of the jackknife-after-bootstrap idea of \citep*{efron1992jackknife,wager2014confidence}--to  construct our jackknife pseudo-values, which are then used to form the EL function for inference. The proposed confidence region retains several desirable features of the conventional EL confidence region: it is range-preserving, transformation-respecting, and its shape is fully determined by the data. For instance, if the parameter of interest has a positive support, the EL confidence region will not include negative values, unlike  Wald-type confidence intervals.

In terms of the asymptotic properties, like many existing inference procedures, such as Wald-tests based on the infinitesimal jackknife or the conventional jackknife, under standard regularity conditions, the proposed jackknife EL procedure attains asymptotically exact size provided that the number of subsamples increases at a sufficiently fast rate; we refer to this regime as dense-subsampling asymptotics. However, when the number of subsamples increases at a slower rate--referred to as sparse-subsampling asymptotics--the limiting distribution of the statistic becomes non-pivotal, leading to size distortions. To address this issue, we further propose a modified EL procedure that not only preserves all the desirable properties of the EL-based inference methods but also remains asymptotically pivotal under both dense-and sparse-subsampling asymptotics.  This modification--analogous to the correction idea for the infinitesimal jackknife variance estimator discussed in Section 4 of \cite*{peng2021bias}--is straightforward to implement. In contrast to the existing jackknife EL-based bias correction methods of \cite*{matsushita2021jackknife,matsushita2024empirical}, our modification does not require leave-more-out estimation and is therefore computationally feasible even under large samples.

Our theoretical results for the proposed EL and modified EL procedures are developed under broad and general assumptions that mirror those considered in \cite*{peng2022rates}. To demonstrate their practical relevance, we further specialize the theory of the modified EL to the honest random forests of \citet{wager2018estimation}, deriving explicit and trackable low-level sufficient conditions. Although we take advantage of the bias bound and variance ratio results for honest trees from \citet{wager2018estimation}, our analysis explicitly accounts for varying degrees of  incompleteness in the $U$-statistic structure due to subsampling, leading to improved finite-sample performance and computational efficiency.

\subsection{Literatures}
Random forests were introduced by \cite{breiman2001random} as a modification of bagging \citep{breiman1996bagging}. Over the past two decades, substantial progress has been made towards understanding their statistical properties. A substantial body of work has established its consistency, convergence rates, and provided insights into the sources of its strong empirical performance; see, for example, \cite*{biau2012analysis}, \cite*{scornet2015consistency}, \cite*{wager2015adaptive},  \cite*{scornet2016asymptotics}, \cite*{mentch2020randomization}, and \cite*{chi2022asymptotic}. See also \cite{scornet2025theory} for a recent survey of the theoretical progress on random forests.

More recently, the asymptotic distributional properties of random-forest point predictions and the associated inferential procedures have attracted considerable attention.
\cite*{wager2014confidence} introduced the infinitesimal jackknife for random forests and their Monte Carlo bias correction methods, providing a computationally tractable variance estimator and clarifying the role of resampling. \cite*{mentch2016quantifying} formally recast  subsampled random forests, after centering, as incomplete $U$-statistics, and derive a central limit theorem for ensemble learners under an incomplete $U$-statistics setup under the assumption that the first order projection of the kernel of $U$-statistics having an asymptotically non-zero variance\footnote{Recent works, such as \cite{diciccio2022clt,peng2022rates}, have noted that this imposes a strong restriction on tree-based learners.}. Under an honesty condition, \citet*{wager2018estimation} decomposed random forest predictions as the sum of a centered infinite-order $U$-statistics and a bias term, establishing a bias bound and pointwise asymptotic normality of forest predictions under explicit and verifiable low-level conditions. Their framework connects random forest predictions to classical $U$-statistic theory while avoiding imposing an asymptotic nonzero variance condition on the first order projection term, enabling asymptotic inference through standard central limit arguments. Within this setup, \citet*{wager2018estimation} further proved the ratio consistency of the infinitesimal jackknife variance estimator, thereby providing a theoretically justified basis for asymptotically valid hypothesis testing and confidence interval construction for random forest predictions.

Subsequent contributions refined these results. In particular, \citet*{peng2022rates} established a central limit theorem and Berry-Esseen bounds for generalized $U$-statistics, explicitly accounting for incompleteness and its impact on the asymptotic distribution. They further examined the variance ratio condition required by their central limit theorem, providing bounds for this ratio for various base learners, such as random potential nearest neighbor estimators and non-adaptive $k$-trees. \cite*{peng2021bias} re-examined the infinitesimal jackknife  as a variance estimator for resampling-based and ensemble methods. They showed that the infinitesimal jackknife can be equivalently derived via ordinary least squares regression, the classical jackknife, or functional derivatives, offering clearer intuition. Furthermore they analyzed its bias, proposed a bias-corrected form, and extended the framework to $U$-statistics, establishing consistency results for both complete and incomplete subsampling schemes relevant to random forests. 
 
Other notably related developments include the high-dimensional Gaussian approximation and bootstrap theory for incomplete infinite-order $U$-statistics established by \cite*{song2019approximating}. \cite*{athey2019generalized} proposed estimators defined by estimating equations localized by random forest weights, together with a corresponding large-sample theory and variance estimator. \cite*{zhou2021v} focused on variance estimation and its relationship with V-statistics, \cite*{ghosal2022infinitesimal} extend the infinitesimal jackknife variance estimator to a wide range of ensemble learning algorithms, demonstrating that it provides a consistent estimator of the covariance between any two model predictions, whereas \cite*{coleman2022scalable} introduced an F-test analogue procedure for efficient hypothesis testing based on permutation-test techniques. \cite*{xu2024variance} derived an unbiased variance estimator for infinite-order $U$-statistics relevant to forest predictions. \cite*{demirkaya2024optimal} investigated ensembles constructed from distributional nearest neighbor learners, establishing their asymptotic distributions, variance estimators, and a simple bias-correction scheme based on a two-scale estimation procedure.  Finally, \cite*{juergens2025jackknife} analyzed the jackknife variance estimator and its variants, providing weaker sufficient conditions for ratio consistency for ensembles employing distributional nearest neighbor base learners, among other recent advances.

We also contribute to the literature on EL (\citealt{Owen1988}; see \citealt{Owen2001Book} for a comprehensive textbook treatment) and related likelihood based inference. Following the seminal work of \cite*{jing2009jackknife}, the jackknife EL and its variants have been widely applied to diverse econometric and statistical problems. For various complete $U$-statistics with a finite and fixed order, \cite{matsushita2021jackknife,matsushita2024empirical}  introduced a modified jackknife EL to correct for the Efron-Stein bias and established its asymptotic pivotality under both conventional and nonstandard asymptotics via leave-more-out bias correction idea from \cite{efron1981jackknife}--see also \cite{chiang2025multiway} for a generalization to multiway clustered data. Under conventional asymptotics, EL methods have been further developed and extended to numerous contexts--see, e.g., \cite*{bertail2006empirical}, \cite*{zhu2006empirical}, \cite*{HjortMcKeagueVan_Keilegom2009AoS}; \cite*{BravoEscancianoVan_Keilegom2020AoS}, and the reviews by \cite{kitamura2006empirical} and \cite*{ChenVan_Keilegom2009Test}.

\noindent\textbf{Notation:} Throughout the paper, for any two nonnegative sequences $(A_n)_{n\in \mathbb N}$ and $(B_n)_{n\in \mathbb N}$, we use $A_{n}\lesssim B_{n}$ to mean that there exists a positive constant $C$ independent of $n$ such that $A_{n}\le CB_{n}$ for all $n$ a.s., $A_{n}\asymp B_{n}$ means $A_{n}\lesssim B_{n}$ and $B_{n}\lesssim A_{n}$, $A_n\ll B_n$ means $A_n/B_n=o(1)$. All the asymptotics are taken as $n\to\infty$. We follow the convention that $0/0=0$ and $\log(\cdot)$ is taken with respect to base $e$.

\section{Empirical likelihood for ensembles}\label{sec:EL}

In this section, we propose and study an EL-based general inference framework that works for generic ensemble learners. We begin by introducing the basic setup. Suppose we observe an independent and identically distributed (i.i.d.) sample $(Z_1, \ldots, Z_n)$ of size $n$ taking values in a measurable space $\mathcal{Z}$. For each sample size $n$, let $s = s_n \leq n$ be a positive integer that may grow with $n$. Denote the collection of all possible index sets corresponding to subsamples of $\{1,...,n\}$ of size $s$ by
\[
I_n = \left\{ \iota = (i_1, \ldots, i_s) : 1 \le i_1 < \cdots < i_s \le n \right\}.
\]
Let $Z_{\iota} = (Z_{i_1} \ldots, Z_{i_s}) \in \mathcal{Z}^{s} $ for $\iota\in I_n$, and $(\omega_{\iota})_{\iota \in I_n}\sim \omega$ be i.i.d. additional  auxiliary randomness that is independent of $(Z_{\iota})_{\iota\in I_n}$. Consider a real-valued function 
\[
h_n: \mathcal{Z}^{s} \times \Omega \to \mathbb{R},
\]
which is assumed to be symmetric in its first $s$ arguments (without loss of generality). We wish to conduct statistical inference on a scalar parameter:
\[
\theta_n = \mathbb{E}\left[h_n(Z_{\iota}, \omega_{\iota})\right]\in\Theta \subset \mathbb{R}.
\]
Throughout the paper, we assume that $\theta_n$ is uniformly bounded and consider an asymptotic regime in which $s \to \infty$ as $n \to \infty$.

We are particularly interested in the setting where $h_n(Z_{\iota}, \omega_{\iota})$ is some estimator constructed using subsample $Z_{\iota}$ of size $s$ and auxiliary randomness $\omega_{\iota}$. In the context of random forests, $h_n(Z_{\iota},\omega_{\iota})$ is considered as a regression or classification tree estimate for $\theta_n$ based on the subsample $Z_\iota$ of size $s$. In Section \ref{sec:RF} below, our general theory is illustrated for tree-based learners. See e.g. \cite{rokach09} for various examples of ensemble learners.

The complete averaging estimator of $\theta_n$ is defined as an infinite-order, generalized (or noisy) $U$-statistic:
\begin{equation}
\check{\theta}_n = \binom{n}{s}^{-1} \sum_{\iota \in I_n} h_n(Z_{\iota}, \omega_{\iota}). \label{eq:U}
\end{equation}
In practice, implementing this estimator can be computationally infeasible when $n$ and $s$ are large. Consequently, it is common to approximate the complete average by computing an average over a randomly chosen subset of $I_n$, yielding the incomplete generalized $U$-statistic:
\begin{equation}
\hat{\theta}_n = \frac{1}{B_n} \sum_{\iota \in I_n} \rho_{\iota} h_n(Z_{\iota}, \omega_{\iota}), \label{eq:IU}
\end{equation}
where $\{\rho_{\iota}\}_{\iota \in I_n}$ is an i.i.d.\ sequence of Bernoulli random variables (independent of $\mathcal{Z}$) indicating which subsamples are selected,  and $B_n = \sum_{\iota \in I_n} \rho_{\iota}$ denotes the total number of selected subsamples.

In the terminology of $U$-statistics, \eqref{eq:U} and \eqref{eq:IU} are referred to as the complete and incomplete $U$-statistics, respectively. The theoretical analysis of the incomplete $U$-statistics is more intricate, primarily because the estimator $\hat{\theta}_n = \hat{\theta}_n(Z_1, \ldots, Z_n,(\omega_\iota)_{\iota\in I_n})$ is not necessarily symmetric in its arguments.

We now construct an EL function for the parameter $\theta_n$. Intuitively, we aim to adapt the jackknife EL approach of \cite{jing2009jackknife}. To illustrate, suppose we were to implement this jackknife EL using the complete $U$-statistic $\check{\theta}_n$ 
defined in \eqref{eq:U}. For each $i=1,\ldots,n$, we may define its 
leave-one-out counterpart as
\begin{equation*}
\check{\theta}_n^{(i)} 
= \binom{n-1}{s}^{-1} 
  \sum_{\iota \in I_n \,:\, i \notin \iota} 
  h_n(Z_{\iota}, \omega_{\iota}).
\end{equation*}
We then consider the corresponding contrasts for each $\theta\in\Theta $
\[
\check S_n(\theta) = \check{\theta}_n - \theta,
\qquad 
\check S_n^{(i)}(\theta) = \check{\theta}_n^{(i)} - \theta,\quad i=1,...,n,
\]
and construct the jackknife pseudo-values
\[
\check V_n^{(i)}(\theta)
= 
n\check S_n(\theta)
- (n-1)\check S_n^{(i)}(\theta),
\]
$i=1,...,n$. By treating the jackknife pseudo values as individual observations, one can form the EL function following the original recipe of \cite{Owen1988}:
\[\check
\ell(\theta)
= -2 \sup_{w_1,\ldots,w_n}
    \sum_{i=1}^n \log(n w_i)
\quad \text{subject to} \quad
w_i \geq 0,\ \sum_{i=1}^n w_i = 1,\ 
\sum_{i=1}^n w_i\, \check V_n^{(i)}(\theta) = 0.
\]
However, as noted in the introduction, a direct application of this jackknife EL is generally infeasible in the settings of random forests or other ensembles. The underlying $U$-statistics are incomplete: computing the complete $U$-statistic is almost always computationally prohibitive, even under moderate sample sizes, and doing so for the corresponding leave-one-out estimators is even less realistic.

To address this issue, we adopt the jackknife-after-subsampling to form our jackknife pseudo-values, which then serve as the basis for constructing the  EL function.
Given a realization of the sampling indicators $\{\rho_\iota\}_{\iota \in I_n}$, let $I_{1n} = \left\{ \iota \in I_n : \rho_\iota = 1 \right\}$ be the collection of subsample indices selected by $\rho_\iota$. For each $i \in \{1, \dots, n\}$, define
\[
I_{1n}^{(i)} = \left\{ \iota \in I_{1n} : i \notin \iota \right\},
\]
that is, $I_{1n}^{(i)}$ consists of all selected subsamples that do not involve the $i$-th observation. 
To conduct inference on $\theta_n$, for each $\theta\in\Theta$, define the contrast $S_n(\theta) = \hat{\theta}_n - \theta$ and its leave-one-out analogue
\[
S_n^{(i)}(\theta) = \frac{1}{B_{n}^{(1)}} \sum_{\iota \in I_{1n}^{(i)}} h_n(Z_\iota, \omega_\iota)-\theta,
\]
where $B_{n}^{(1)}=\frac{n-s}{n}B_n$. Then for $i=1,\ldots,n$, the jackknife pseudo-values are given by
\[
V_n^{(i)}(\theta) = n S_n(\theta) - (n-1) S_n^{(i)}(\theta).
\]
Observe that the proposed jackknife pseudo-values are computationally straightforward, as they depend only on the $B_n$ regression trees that have already been constructed for estimating the random forest $\hat \theta_n$. Moreover, the denominator term $B_n^{(1)}$ in $S_n^{(i)}$ does not involve heterogeneous sample sizes and is identical for all $S_n^{(i)}$, $i = 1,\ldots,n$.

Using these pseudo-values as estimating equations, our EL function for $\theta_n$ is constructed as
\[
\ell(\theta) = -2 \sup_{w_1, \ldots, w_n} \sum_{i=1}^n \log(n w_i)
\quad \text{subject to} \quad
w_i \geq 0, \quad \sum_{i=1}^n w_i = 1, \quad \sum_{i=1}^n w_i V_n^{(i)}(\theta) = 0.
\]
Although this optimization involves $n$ weights $\{w_i\}$, its dual form can be derived via the method of Lagrange multipliers. The resulting expression is:
\begin{equation} \label{eq:JEL}
\ell(\theta) = 2 \sup_{\lambda\in \mathbb{R}} \sum_{i=1}^n \log (1 + \lambda V_n^{(i)}(\theta)), 
\end{equation}
where $\lambda$ is the Lagrange multiplier associated with the estimating equation constraint. Algorithm \ref{alg:EL} below summarizes its implementation.

\begin{algorithm}[htbp]
\caption{Empirical Likelihood (EL) for Ensemble Learners}
\label{alg:EL}
\begin{algorithmic}[1]
\REQUIRE  training set $(Z_1,\ldots,Z_n)$, subsample size $s$, ensemble learner $h_n$, candidate value $\theta \in \Theta$
\STATE Generate an ensemble of trees (or base learners) using subsampling:
       for each $b=1,\ldots,B_n$, draw a subsample index set $\iota_b \in I_n$
       of size $s$ and build a tree with prediction
       $h_b := h_n(Z_{\iota_b},\omega_{\iota_b})$.
\STATE Compute the incomplete generalised $U$-statistic (ensemble prediction)
       \[
         \hat{\theta}_n \;=\; \frac{1}{B_n} \sum_{b=1}^{B_n} h_b .
       \]
\STATE For each $i=1,\ldots,n$, define the index set of trees that do not use
       observation $i$:
       \[
         I^{(i)}_{1n} \;:=\; \{\, b \in \{1,\ldots,B_n\} : i \notin \iota_b \,\},
       \]
       and let $B_n^{(1)} := |I^{(i)}_{1n}|$ (identical for all $i$ under the
       jackknife-after-subsampling scheme).
\STATE For the given $\theta$, compute the contrasts
       \[
         S_n(\theta) := \hat{\theta}_n - \theta,\qquad
         S_n^{(i)}(\theta) := \frac{1}{B_n^{(1)}} \sum_{b \in I^{(i)}_{1n}} h_b - \theta ,
       \]
       for $i=1,\ldots,n$.
\STATE Construct the jackknife pseudo-values
       \[
         V_n^{(i)}(\theta) := n S_n(\theta) - (n-1) S_n^{(i)}(\theta),
         \qquad i=1,\ldots,n.
       \]
\STATE Obtain the Lagrange multiplier $\hat{\lambda}(\theta)$ by solving the
       one-dimensional equation
       \[
         \sum_{i=1}^n
         \frac{V_n^{(i)}(\theta)}{1 + \lambda\, V_n^{(i)}(\theta)} = 0
       \]
       for $\lambda$ (e.g., by a numerical root-finding method).
\STATE Compute the empirical likelihood (EL) statistic
       \[
         \ell(\theta) :=
         2 \sum_{i=1}^n \log\!\bigl(1 + \hat{\lambda}(\theta)\,
                                     V_n^{(i)}(\theta)\bigr).
       \]
\ENSURE EL statistic $\ell(\theta)$; an approximate $100(1-\alpha)\%$ confidence
        region for $\theta_n$ is
        \[
          \mathcal{C}_{\mathrm{EL}} :=
          \bigl\{ \theta \in \Theta : \ell(\theta) \le \chi^2_{1,\alpha} \bigr\}.
        \]
\end{algorithmic}
\end{algorithm}

Hereafter, for notational convenience, we recenter the kernel so that $\theta_n = 0$. 
Denote
\(
\zeta_{n,s} = \mathbb{V}\left(h_n(Z_{1},\ldots,Z_{s},\omega)\right)
\)
for  the variance of the kernel, and
\(
\zeta_{n,1} 
= \mathbb{V}\left(\mathbb{E}\big[h_n(Z_{1},\ldots,Z_{s},\omega)\mid Z_1\big]\right),
\)
the variance of its first--order projection. To analyze the asymptotic behavior of the jackknife EL statistic $\ell(\theta_n)$, we impose the following assumptions.
 
\begin{assumption}\label{asm:JEL}
Consider the setup of this section and assume that
\begin{description}
    \item[(i)] there exist positive constants $C$ and $c$ such that $\mathbb{E}[|h_n(Z_{\iota},\omega)|^{6}]\le C < \infty$ and $\zeta_{n,s} \ge c > 0$ for all $n$;
    \item[(ii)] $\max_{\iota \in I_n}|h_n(Z_\iota,\omega)|=o_p((n\zeta_{n,1})^{1/2})$;
     \item[(iii)] as $n\to\infty$, it holds $s\to\infty$, $s/n \to 0$, and
\begin{equation} \label{eq:zs1}
\frac{s}{n}\frac{\zeta_{n,s}}{s\zeta_{n,1}} =o(1).
\end{equation}
\end{description}
\end{assumption}
Assumption \ref{asm:JEL} (i) is on moments of the $U$-statistic kernel $h_n(Z_\iota,\omega)$. The boundedness condition in Assumption \ref{asm:JEL} (ii) is imposed to derive a quadratic expansion for the EL statistic. Assumption \ref{asm:JEL} (iii) is on our asymptotic framework. The subsample size $s$ grows at a slower rate than $n$, and the variance ratio $\zeta_{n,s}/s\zeta_{n,1}$ needs to be controlled as in (\ref{eq:zs1}). The variance ratio condition in (\ref{eq:zs1}) is required to apply the central limit theorem for generalized $U$-statistics by \citet*[Theorem 2]{peng2022rates}.

Define the expected subsample size as 
\(
N_n = \mathbb{E}[B_n] = p_n |I_n|\), where
\(p_n = \mathbb{P}\{\rho_{\iota} = 1\}
\).
Since under the current setup, following Bernstein's inequality, \(B_n\) concentrates rapidly around \(N_n\) as \(n\) grows, all asymptotic behavior may be characterized in terms of \(N_n\) in place of \(B_n\).
  We call the asymptotic regime to be under dense-subsampling asymptotics if 
\begin{equation} \label{eq:dense}
\frac{n}{N_n}\frac{\zeta_{n,s}}{s\zeta_{n,1}} =o(1),
\end{equation}
and under sparse-subsampling asymptotics if
\begin{equation}
\frac{n}{N_n}\frac{\zeta_{n,s}}{s\zeta_{n,1}} \to C_* \in (0,\infty).\label{eq:sparse}
\end{equation}
Intuitively, dense-subsampling asymptotics arises when \(N_n\) increases sufficiently fast relative to \(n\), whereas sparse-subsampling asymptotics corresponds to settings in which \(N_n\) diverges, but at a slower rate. See Section \ref{sec:mEL} for a detailed discussion on the two asymptotic regimes.

The  dense-subsampling asymptotics is a crucial condition in the existing variance estimation methods that take into account of the incompleteness of the $U$-statistic structure. For example, a similar condition is employed in \citet[Theorem 4]{peng2021bias} to obtain consistency of the infinitesimal jackknife variance estimator.

Under these assumptions, one can establish asymptotic pivotality of the EL statistic.
\begin{theorem} \label{thm:JEL}
Suppose that Assumption \ref{asm:JEL} holds true. 
Then under dense-subsampling asymptotics in \eqref{eq:dense}, we have
\[
\ell(\theta_n)\overset{d}{\to}\chi_{1}^{2}
\]
\end{theorem}
For a proof, see Section \ref{app:proof:thm:JEL} in the Appendix.

Based on this theorem, for $\alpha\in[0,1]$, the $100(1-\alpha)\%$ asymptotic confidence region for $\theta_n$ can be constructed as $\{\theta : \ell(\theta) \le \chi^2_{1,\alpha}\}$, where $\chi^2_{1,\alpha}$ is the $(1-\alpha)$-th quantile of the $\chi^2_1$ distribution. This  EL  confidence region shares desirable properties with the conventional EL confidence region, such as  being transformation respecting and having a data decided shape for the confidence region. In addition, it is range preserving in the sense that, if the researcher knows that $\theta_n \ge 0$ (e.g., for a variance or a probability), the EL confidence region respects this constraint and excludes negative values. On the other hand, the Wald-type confidence regions (i.e., $\hat{\theta}_n\pm 1.96\cdot$se) may contain negative values in finite samples. Also the EL confidence region can be asymmetric around the point estimator $\hat{\theta}_n$ whose shape is determined by data. For a more detailed account of these properties of EL confidence regions, see \citet[Section 3]{Owen1988}.

The dense-subsampling asymptotic condition in (\ref{eq:dense}), which requires $N_n$ to grow sufficiently fast relative to the variance ratio $\zeta_{n,s}/\zeta_{n,1}$, is crucial for ensuring the asymptotic pivotality of $\ell(\theta_n)$. However, satisfying this condition can be computationally demanding or even infeasible in large samples, as $N_n$ must increase at a faster rate with $n$. In practice, moderate violations are common, and inference procedures that rely on this condition often exhibit over-coverage in finite-samples. This issue is not specific to our EL procedure; other variance estimation approaches, such as the jackknife and infinitesimal jackknife methods 
are also prone to similar finite sample distortions.

\section{Alternative asymptotics and modified empirical likelihood}\label{sec:mEL}
As   previously discussed, the EL-based procedure introduced in Section \ref{sec:EL} can be overly conservative in finite samples.
In this section, we propose and analyze a modified EL inference procedure for generic ensemble learners by first introducing an alternative asymptotic framework to understand the over-coverage property of the EL statistic, and then develop a modified statistic to achieve better coverage. 

In order to grasp the intuition behind the over-coverage problem, it is insightful to consider Hoeffding's decomposition for the generalized $U$-statistics following \cite*{peng2022rates}:
\begin{align}
\frac{1}{{n \choose s}}\sum_{\iota\in I_{n}}\frac{\rho_{\iota}}{p_{n}}h_n(Z_{\iota},\omega_{\iota}) & = \sum_{j=1}^{s}{s \choose j}\left\{ {n \choose j}^{-1}\sum_{(n,j)}h^{(j)}_n(Z_{i_{1}},\ldots,Z_{i_{j}},\omega_{\iota})\right\} \nonumber \\ 
 & \quad +{n \choose s}^{-1}\sum_{(n,s)}\left(\frac{\rho_{\iota}}{p_{n}}-1\right)h_n(Z_{\iota_{1}},\ldots,Z_{\iota_{s}},\omega_{\iota})\nonumber \\
 & =: U_{n}+\tilde{U}_{n}, \label{eq:Hoeffding}
\end{align}
where $(n,j)=\{(i_{1},\ldots,i_{j}):1\le i_{1}<\ldots<i_{j}\le n\}$ and $h^{(j)}_n(Z_{i_{1}},\ldots,Z_{i_{j}},\omega_{\iota})$ is the $U$-statistic kernel for $j$-th order Hoeffding projection--see Equation \eqref{eq:hj} in the Appendix for its formal definition. Note that the term $\tilde{U}_{n}$ is due to incompleteness of the $U$-statistic. The key condition in (\ref{eq:dense}) guarantees that $\tilde{U}_{n}$ is asymptotically negligible, which corresponds to the dense-subsampling asymptotics-- for mathematical details, see Lemma \ref{lem:A1_standard} in the Appendix. However, in practice, the expected number of ensembles $N_n$ may be relatively small so that \eqref{eq:dense} is less plausible and the contribution from $\tilde{U}_{n}$ can be non-negligible. To achieve better approximation in this scenario, we consider the alternative asymptotic framework where both $U_{n}$ and $\tilde{U}_{n}$ match at the same order and modify the EL statistic to accommodate the contribution associated with $\tilde{U}_{n}$, that is, the sparse-subsampling asymptotics defined in (\ref{eq:sparse}).

We propose the following modification to our jackknife pseudo values for the EL statistic
\begin{equation}
\tilde{V}_{n}^{(i)}(\theta)=V_{n}^{(i)}(\hat{\theta}_{n})-\sqrt{\frac{\hat{\mathcal{V}}_{1}}{\hat{\mathcal{V}}_{1}+\hat{\mathcal{V}}_{2}}}\{V_{n}^{(i)}(\hat{\theta}_{n})-V_{n}^{(i)}(\theta)\}, \label{eq:V_tilde}
\end{equation}
where $\hat{\mathcal{V}}_{1}=\frac{1}{n}\sum_{i=1}^{n}V_{n}^{(i)}(\hat{\theta}_{n})^{2}$, $\hat{\mathcal{V}}_{2}=\frac{n(s-1)}{B_{n}}\hat{\zeta}_{n,s}$, and
\[
\hat{\zeta}_{n,s}=\frac{1}{B_n}\sum_{\iota \in I_{1n}}  h_n(Z_\iota,\omega_\iota)^2-\hat\theta_n^2
\]
is the sample counterpart of $\zeta_{n,s}$. The term $\hat{\mathcal{V}}_{2}$ is an additional component to internalize the influence from the term $\tilde{U}_{n}$ in \eqref{eq:Hoeffding}. Based on this modified jackknife pseudo values, our modified EL function can be defined as
\begin{equation}
\tilde{\ell}(\theta)=2\sup_{\lambda}\sum_{i=1}^{n}\log(1+\lambda\tilde{V}_{n}^{(i)}(\theta)).\label{eq:mJEL}
\end{equation} 
An implementation is summarized in Algorithm \ref{alg:mEL} below.

\begin{algorithm}[htbp]
\caption{modified Empirical Likelihood (mEL) for Ensemble Learners}
\label{alg:mEL}
\begin{algorithmic}[1]
\REQUIRE training set $(Z_1,\ldots,Z_n)$, subsample size $s$, ensemble learner $h_n$, candidate value $\theta \in \Theta$
\STATE Using the same ensemble as in Algorithm~\ref{alg:EL}, compute
       $\hat{\theta}_n = B_n^{-1} \sum_{b=1}^{B_n} h_b$.
\STATE For each $i=1,\ldots,n$, define $I^{(i)}_{1n}$ and $B_n^{(1)}$ as in
       Algorithm~\ref{alg:EL} and compute
       \[
         S_n(\theta) := \hat{\theta}_n - \theta,\qquad
         S_n^{(i)}(\theta) := \frac{1}{B_n^{(1)}} \sum_{b \in I^{(i)}_{1n}} h_b - \theta .
       \]
\STATE Construct jackknife pseudo-values (at a generic $\theta$)
       \[
         V_n^{(i)}(\theta) := n S_n(\theta) - (n-1) S_n^{(i)}(\theta),
         \qquad i=1,\ldots,n.
       \]
\STATE Evaluate the pseudo-values at $\hat{\theta}_n$ and form
       \[
         \widehat{\mathcal V}_1 :=
         \frac{1}{n} \sum_{i=1}^n V_n^{(i)}(\hat{\theta}_n)^2.
       \]
\STATE Estimate the kernel variance components
       \[
         \widehat{\zeta}_{n,s}
         := \frac{1}{B_n} \sum_{b=1}^{B_n}
            \bigl(h_b^2\bigr) - \hat{\theta}_n^{\,2},
     \qquad
         \widehat{\mathcal V}_2 :=
         \frac{n(s-1)}{B_n}\, \widehat{\zeta}_{n,s}.
       \]
\STATE For the given $\theta$, define the modified pseudo-values
       \[
         \widetilde{V}_n^{(i)}(\theta)
         :=
         V_n^{(i)}(\hat{\theta}_n)
         -  \sqrt{\frac{\widehat{\mathcal V}_1}{\widehat{\mathcal V}_1 + \widehat{\mathcal V}_2}}.\Bigl\{ V_n^{(i)}(\hat{\theta}_n) - V_n^{(i)}(\theta) \Bigr\},
         \qquad i=1,\ldots,n.
       \]
\STATE Obtain the Lagrange multiplier $\widetilde{\lambda}(\theta)$ by solving for $\lambda$
       \[
         \sum_{i=1}^n
         \frac{\widetilde{V}_n^{(i)}(\theta)}{1 + \lambda\, \widetilde{V}_n^{(i)}(\theta)} = 0.
       \]
       
\STATE Compute the modified empirical likelihood statistic
       \[
         \widetilde{\ell}(\theta) :=
         2 \sum_{i=1}^n
           \log\!\bigl(1 + \widetilde{\lambda}(\theta)\,
                           \widetilde{V}_n^{(i)}(\theta)\bigr).
       \]
\ENSURE mEL statistic $\widetilde{\ell}(\theta)$; an approximate
        $100(1-\alpha)\%$ confidence region for $\theta_n$ is
        \[
          \mathcal{C}_{\mathrm{mEL}} :=
          \bigl\{ \theta \in \Theta : \widetilde{\ell}(\theta)
                                     \le \chi^2_{1,\alpha} \bigr\}.
        \]
\end{algorithmic}
\end{algorithm}

To study the property of our inference procedure under the sparse-subsampling asymptotics in (\ref{eq:sparse}), we add the following assumption.
\begin{assumption}\label{asm:mJEL}
$\|\E[h_n(Z_1,...,Z_s,\omega)^2|Z_1]\|_\infty=o\left(\frac{n}{s}\right)$ a.s.
\end{assumption}
Intuitively this assumption is used to control the covariance of $U_n$ and $\tilde{U}_n$ in \eqref{eq:Hoeffding}--see Lemma \ref{lem:A1_nonstandard} in the Appendix for details. Under this assumption, the asymptotic distributions  of the EL and modified EL statistics under both dense- and sparse-subsampling asymptotics are obtained as follows.
\begin{theorem} \label{thm:mJEL}
Suppose that Assumptions \ref{asm:JEL} and \ref{asm:mJEL} hold true, and as $n\to\infty$, it holds $N_n\to\infty$. Then under both dense- and sparse-subsampling asymptotics, i.e.
\begin{equation}
\frac{n}{N_n}\frac{\zeta_{n,s}}{s\zeta_{n,1}} \to C_* \in [0,\infty),
\end{equation}
 we have
\begin{align*}
    {\rm (i):}\quad & \ell(\theta_n)\overset{d}{\to} (1+C_*)^{-1}\chi_{1}^{2}, \\
    {\rm (ii):}\quad & \tilde{\ell}(\theta_n)\overset{d}{\to}\chi_{1}^{2}.
\end{align*}
\end{theorem}

For a proof, see Sections \ref{app:proof:thm:mJEL(i)} and \ref{app:proof:thm:mJEL(ii)} in the Appendix.

Since $C_*\ge0$, if we use the $\chi_1^2$ critical value for the EL statistic $\ell(\theta)$, the resulting EL confidence region $\{\theta:\ell(\theta)\le\chi_{1,\alpha}^{2}\}$ will exhibit over-coverage. On the other hand, the modified EL statistic $\tilde{\ell}(\theta_n)$ restores asymptotic pivotality and admits the calibration by the $\chi_1^2$ critical value.

The modification in \eqref{eq:V_tilde} targets the \cite{efron1981jackknife}-type bias of the jackknife-after-subsampling variance, which overestimates the contribution of the highest-order term by $s-1$. For genuine finite-order $U$-statistics with fixed $s$, this higher-order bias is asymptotically negligible; however, in random-forest asymptotics where $s\to\infty$, the inflation can become non-negligible and even substantial under sparse-subsampling regimes. Similar Monte Carlo bias corrections are also proposed in \citet[Equation (8)]{wager2014confidence} for bagging, but they focus on the variance estimation rather than distributional approximation.


At first glance, this adjustment may also appear similar to the modification of \citet{matsushita2021jackknife} for finite-order $U$-statistics, as both approaches alter the curvature of the jackknife pseudo-values around $\hat{\theta}_n$. Yet the underlying mechanisms differ markedly. In several settings considered by \citet{matsushita2021jackknife}--such as link-formation probability estimation in sparse networks--the second-order Hoeffding projection remains asymptotically normal and of the same order as the H\'ajek projection, while the jackknife variance overestimates the variance of this second order component by a factor of $2$ \citep*{efron1981jackknife}. Their correction therefore focuses on this double-counting, implemented through leave-two-out estimators. In contrast, under the CLT condition in \eqref{eq:zs1}, all of our Hoeffding projections of orders $2,\dots,s-1$, and the corresponding jackknife variance components, are asymptotically negligible under both dense- and sparse-subsampling asymptotics.\footnote{Moreover, these higher-order projections need not be asymptotically normal, so a na\"ive adaptation of the \citet{matsushita2021jackknife} modification would lack theoretical justification in the present setup, aside from imposing a substantial computational burden.} Under the sparse subsampling asymptotics,  the $s$-th order projection remains asymptotically normal, but its contribution is overestimated by a factor of $s$, an inflation that can be severe when $s$ is large. Our modification directly corrects this jackknife bias without requiring leave-more-out estimators.

\section{Application to honest random forests} \label{sec:RF}

Thus far, the theoretical results for both the EL and modified EL procedures have been developed under high-level, general conditions applicable to generic ensemble learners. In this section, we derive low-level sufficient conditions specialized to the honest random forests of \citet{wager2018estimation} and provide validity for modified EL-based inference.

We first state a useful result for generic tree-based learners. For $h_n$ consists of tree-based learners, it is shown in Lemma \ref{lem:maxV_sufficient} below that Assumption \ref{asm:lem:maxV_sufficient} below is a sufficient condition for both Assumptions \ref{asm:JEL} (ii) and \ref{asm:mJEL}.

\begin{assumption}\label{asm:lem:maxV_sufficient}  
Assume that 
\(
\zeta_{n,1}\ge C/n^{1-2/q}
\)
for some constant $C>0$ independent of $n$, $\mathbb{E}[|Y_{1}|^{q}]<\infty$ for a $q>4$, and the kernel $h_n$ takes the form of $h_n(Z_{1},...,Z_{s},\omega_{1...s})=\sum_{i=1}^{s}S_{i}Y_{i}$
with some $S_i=S_i(Z_{1},...,Z_{s},\omega_{1...s})$ such that $\sum_{i=1}^{s}S_{i}=1$ and $\frac{1}{2k-1}\le S_{i}\mathbb{I}\{S_{i}\ne0\}\le\frac{1}{k}$
for a fixed $k$.
\end{assumption} 

\begin{lemma}\label{lem:maxV_sufficient}
Under Assumption \ref{asm:lem:maxV_sufficient}, it holds that \(\max_{\iota\in I_n}|h_n(Z_\iota,\omega)|=o((n\zeta_{n,1})^{1/2})\) a.s. In addition, if $s\lesssim n^{1-2/q}$, then it holds $\|\E[h_n(Z_1,...,Z_s,\omega)^2|Z_1]\|_\infty=o\left(n/s\right)$ a.s.
\end{lemma}

For a proof, see Section \ref{app:proof:lem:maxV_sufficient} in the Appendix.

The following defines the honesty condition for trees as introduced in \citet{wager2018estimation}.

\begin{assumption}[Honesty]
A tree trained on a training sample $\{Z_1,...,Z_s\}$ is \emph{honest} if for each $i\in\{1,...,s\}$, it only uses the response $Y_i$ to predict the within-node outcome or to decide where to place the splits, but not both.
\end{assumption}
\cite{wager2018estimation} provide examples of honest trees. One popular example is the double-sample tree, which achieves honesty by randomly dividing its training subsample into two halves $\mathcal I$ and $\mathcal J$. Then it uses $\mathcal J$-sample to place the tree splits while holding out the $\mathcal I$-sample to do within-node prediction.
The following provides a low-level sufficient condition for asymptotic theory for modified EL implemented with honest random forests.
\begin{theorem}[modified EL for honest random forest]\label{thm:mJEL-honestRF}
Suppose that we observe an i.i.d. training example $Z_i = (X_i, Y_i) \in [0, 1]^d \times \mathbb{R}$ with $\E[|Y|^q]<\infty$ for some $q>6$. Further assume that the features satisfy  $X_i \sim \mathrm{Unif}([0, 1]^d)$, $ \mu(x)=\E[Y | X = x]$ and $ \E[Y^2 | X = x]$ are Lipschitz continuous, $\Var(Y |X = x) > 0$, and $\E[|Y - \mu(x)|^{2} | X = x] \leq M$ for some $M > 0$, uniformly over $x \in [0, 1]^d$.

Define the following properties for a regression tree.
    
\begin{enumerate}
\item {\bf Random-split} in the sense that, at every step of the tree-growing procedure, marginalizing over the auxiliary randomness $\omega$, the probability that the next split occurs along the $j$-th feature is bounded below by $p/d$ for some $0 < p \leq 1$, for all $j = 1, \ldots, d$.
\item {\bf $\alpha$-regular} in the sense that each split leaves at least a fraction $\alpha$ of the available training examples on each side of the split and, moreover, the trees are fully grown to depth $k$ for some fixed $k \in \mathbb{N}$, i.e., there are between $k$ and $2k - 1$ observations in each terminal node of the tree.
\item {\bf Symmetric} in the sense that the (possibly randomized) output of the predictor does not depend on the order ($i = 1, 2, ...$) in which the training data are indexed.
\end{enumerate}
For any given $x_0\in [0,1]^d$,
let $h_n$ denote an honest, $\alpha$-regular ($\alpha \leq 0.2$), symmetric, and random-split tree based on a sample of size $s$, estimating $\theta_0 = \mu(x_0)$. Let $\hat{\theta}_n$ be the estimator of $\theta_0$ obtained from a random forest constructed using these regression trees, each trained on a subsample of size $s$. Finally, suppose that 
\(
s =s_n\asymp n^{\beta}
\)
for some 
\[
\beta_{\min} := 1 - 
\left(
1 + 
\frac{d}{p}
\frac{\log(\alpha^{-1})}{\log((1 - \alpha)^{-1})}
\right)^{-1}
< \beta < 1-\frac{2}{q}.
\]
Then Assumptions \ref{asm:JEL}-\ref{asm:lem:maxV_sufficient} are satisfied. Therefore, under both dense-subsampling and sparse-subsampling asymptotics, it holds
\[
\tilde{\ell}(\theta_0)\overset{d}{\to}\chi_{1}^{2}.
\]
\end{theorem}

For a proof, see Section \ref{app:proof:thm:mJEL-honestRF} in the Appendix. 

Typically, the object of interest is the conditional expectation rather than the mean of the regression trees. In such a case, we may use $\theta_0 = \mu(x_0)$ in place of $\theta_n$ in the modified EL function $\tilde \ell(\cdot)$, owing to the bias bound established in \cite{wager2018estimation}.

Although Theorem \ref{thm:mJEL-honestRF} builds on our Theorem \ref{thm:mJEL}, we benefited from several theoretical properties of honest trees established in \cite{wager2018estimation}. The regularity conditions imposed are nearly identical to those in \citet[Theorem 1]{wager2018estimation}, except that our Theorem \ref{thm:mJEL-honestRF} additionally requires the existence of a higher-order \( q \)-th moment and imposes a slightly stronger restriction \( \beta < 1 - \frac{2}{q} \) in place of their condition \( \beta < 1 \). Moreover, unlike their analysis, which is based on complete $U$-statistics, our result explicitly accommodates the incompleteness of random forests and quantifies the effect of varying degrees of subsampling-induced incompleteness.

It is also worth noting that, under the conditions of Theorem \ref{thm:mJEL-honestRF}, \citet[Theorem 5]{wager2018estimation} yields
\begin{align}
    \frac{n}{N_n}\frac{\zeta_{n,s}}{s\zeta_{n,1}}
    \lesssim
    \frac{n}{N_n}(\log s)^d  \le
    \frac{n}{N_n}(\log n)^d.
    \label{eq:n-N_n-ratio}
\end{align}
Thus, at least theoretically, dense--subsampling asymptotics can be ensured by taking \(N_n \gg n(\log n)^d\) regardless of the order of $s$. In finite samples, however, the left-hand side of \eqref{eq:n-N_n-ratio} may remain non-negligible even when \(N_n\) satisfies this growth condition. For this reason, we recommend using modified EL over EL to conduct statistical inference in practice.

\section{Simulation studies}

We conduct a series of simulation studies to evaluate the finite-sample performance of the proposed method under both linear and non-linear data-generating processes (DGPs). Throughout the simulations, we generate $d = 6$ predictors independently as each $X_j \sim \mathrm{Unif}(0,1)$, and assess pointwise inference accuracy at the representative evaluation point $x_0 \in \{0.4, \ldots, 0.4\}$. The sample size $n$ and subsample size $s$ vary across designs to examine the effect of dimensionality and subsampling density on estimation and coverage. All results are averaged over $2,000$ replications.

The first DGP, denoted \textit{MLR}, represents a well-specified linear model:
\[
Y = 2X_1 + 3X_2 - 5X_3 - X_4 + 1 + 2\varepsilon, \qquad \varepsilon \sim \mathcal{N}(0,1),
\]
with the errors generated independently from the predictors. This design provides a baseline where the model structure aligns with the common linear models, allowing us to benchmark the performance in the absence of more complex nonlinearity.

To investigate robustness in more nonlinear settings, we consider a non-linear design, denoted \textit{MARS}, following the spirit of \citet{friedman1991multivariate}:
\[
Y = \sin\!\big(\pi X_1 X_2\big) + 2\,(X_3 - 0.05)^2 - X_4 + 0.5 X_5 + 2\varepsilon, \qquad \varepsilon \sim \mathcal{N}(0,1),
\]
with the errors generated independently from the predictors.
This design introduces strong interactions and local nonlinearity while maintaining moderate noise, providing a challenging environment for evaluating both flexibility and inferential validity. In both scenarios, $X_6$ is excluded from the DGPs. Together, the \textit{MLR} and \textit{MARS} settings span two representative regimes--parametric and non-parametric--highlighting how the proposed inference procedure adapts across smoothness and structural complexity.

Random forest predictions at $x_0$ are obtained using the \texttt{grf} regression forest, with $N_n = \mathbb{E}[B_n] = \lceil (10n)^{\alpha} \rceil$, where $\alpha$ controls the effective number of trees. Each tree is trained on a subsample of size $s = \lceil n^{0.8} \rceil$, allowing us to examine the effect of the forest size $N_n$ on inference accuracy. During tree construction, the number of features considered at each split is set to \texttt{mtry} $= d/2$ and hence the estimation and prediction correspond to random forests \citep{breiman2001random} rather than bagging \citep{breiman1996bagging}. To minimize the bias, which is important for statistical inference and hypothesis testing, the terminal node size is fixed at \texttt{min.node.size} $=1$, and the honesty option is enabled (\texttt{honesty} $=$ \texttt{TRUE}). All hyperparameters are held constant across competing methods to ensure that differences in performance reflect only inferential, rather than tuning, variation.

Inference procedures considered in the simulations include the following four alternatives: 
(i) Wald-test based on the infinitesimal jackknife (IJ) variance for random forests, computed via the \texttt{predict} function with 
\texttt{estimate.variance} $=$ \texttt{TRUE}; 
(ii) Wald-test based on leave-one-out jackknife variance $V_J$; 
(iii) test based on the (unmodified) empirical likelihood (EL) proposed in Section \ref{sec:EL}; and 
(iv) the modified empirical likelihood (mEL) proposed in \ref{sec:mEL}.
Simulation results are reported for sample sizes $n = 200, 400, 800$, under two regimes for the number of trees: \emph{Large} $N$ with $\alpha = 1.2$ and \emph{Small} $N$ with $\alpha = 1.1$. These designs are intended to represent dense-subsampling asymptotics and sparse-subsampling asymptotics, respectively, and allow us to examine how the choice of $N_n$ affects the finite-sample behavior of the corresponding inference procedures.

Across all DGPs, the Wald test based on IJ variance exhibits notable size distortion in finite samples. In our simulation design, this oversize problem persists even as the sample size increases. By contrast, the jackknife variance-based Wald test and the (unmodified) EL procedure both achieve valid, though conservative, size control--particularly under the small-$N$ configurations. Overall, the modified EL  delivers the most stable finite sample performance across designs and sample sizes, attaining coverage rates close to the nominal level even in sparse-subsampling regimes.

\newpage
\begin{figure}[!ht]
    \centering
    \includegraphics[width=\textwidth]{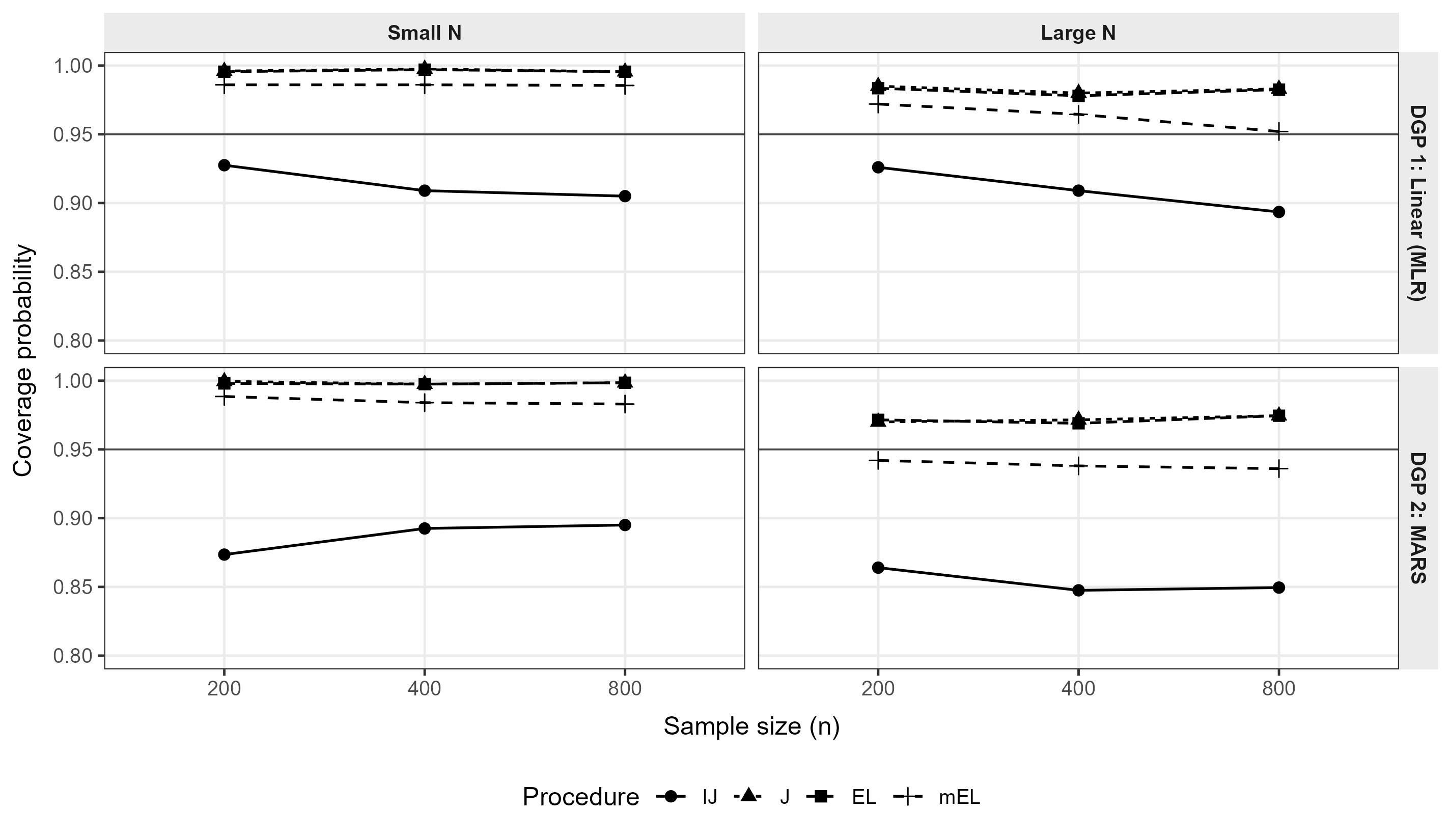}
    \caption{Coverage probabilities of 95\% confidence intervals across sample sizes $n \in \{200,400,800\}$, 
             Linear (MLR) and nonlinear (MARS) DGPs, and four procedures (IJ, J, EL, mEL). The horizontal line marks the nominal 0.95 level.}
    \label{fig:coverage_1}
\end{figure}
\newpage
\appendix

\section{Proof of main results}
\subsection{Hoeffding's decomposition}

Denote $\zeta_{n,c}=\mathbb{V}(h_{n,c}(Z_{1},\ldots,Z_{c}))$, where
\[
h_{n,c}(z_{1},\ldots,z_{c}) =\mathbb{E}[h_{n}(z_{1},\ldots,z_{c},Z_{c+1},\ldots,Z_{s},\omega)]-\theta_n,
\]
for $c=1,\ldots,s-1$. Furthermore,
let $h_n^{(1)}(x_{1})=h_{n,1}(x_{1})$ and define recursively 
\[
h_n^{(c)}(z_{1},\ldots,z_{c})=h_{n,c}(z_{1},\ldots,z_{c})-\sum_{j=1}^{c-1}\sum_{(c,j)}h_n^{(j)}(z_{i_{1}},\ldots,z_{i_{j}}),
\]
for $c=2,\ldots,s-1$, where $(c,j)=\{(i_{1},\ldots,i_{j}):1\le i_{1}<\ldots<i_{j}\le c\}$.
Also define
\begin{equation}
h_n^{(s)}(z_{1},\ldots,z_{s},\omega)=h_n(z_{1},\ldots,z_{s},\omega)-\sum_{j=1}^{s-1}\sum_{(s,j)}h_n^{(j)}(z_{i_{1}},\ldots,z_{i_{j}}). \label{eq:hj}
\end{equation}
Hereafter, when no ambiguity arises, we suppress the subscript $n$ and denote $\theta=\theta_n$, $\hat{\theta}=\hat{\theta}_n$, $\zeta_1=\zeta_{n,1}$, and $\zeta_s=\zeta_{n,s}$.

\subsection{Proof for Theorem \ref{thm:JEL}}\label{app:proof:thm:JEL}
Due to complication coming from the various trajectories of the orders of $s$ and $\zeta_1$, instead of following the standard asymptotic arguments for EL as in, e.g. \cite{Owen1990}, we 
define $\sigma_{n}=\sqrt{s^{2}\zeta_{1}}$ and reparametrize the EL function in Equation \eqref{eq:JEL} by $\ell(\cdot)=2\sup_{\gamma}\sum_{i=1}^{n}\log(1+\gamma\sigma_{n}^{-1}V_{n}^{(i)}(\cdot))$. By Lemmas \ref{lem:V}-\ref{lem:maxV} and the argument of \citet[][around Equation (2.14)]{Owen1990}, it follows that $$\hat{\gamma}=\arg\max_{\gamma}\sum_{i=1}^{n}\log(1+\gamma\sigma_{n}^{-1}V_{n}^{(i)}(\theta))=O_{p}(n^{-1/2}).$$

Next, we obtain an asymptotic approximation for $\hat{\gamma}$. The
first-order condition for $\hat{\gamma}$ satisfies
\[
    0=\frac{1}{n}\sum_{i=1}^{n}\frac{\sigma_{n}^{-1}V_{n}^{(i)}(\theta)}{1+\hat{\gamma}\sigma_{n}^{-1}V_{n}^{(i)}(\theta)}=\frac{1}{\sigma_{n}n}\sum_{i=1}^{n}V_{n}^{(i)}(\theta)-\frac{1}{\sigma_{n}^{2}n}\sum_{i=1}^{n}V_{n}^{(i)}(\theta)^{2}\hat{\gamma}+\frac{1}{n}\sum_{i=1}^{n}\frac{\sigma_{n}^{-3}V_{n}^{(i)}(\theta)^{3}\hat{\gamma}^{2}}{1+\hat{\gamma}\sigma_{n}^{-1}V_{n}^{(i)}(\theta)},
\]
where the second equality follows from the identity $(1+x)^{-1}=1-x+x^{2}(1+x)^{-1}$.
By Lemmas \ref{lem:VV_standard} and \ref{lem:maxV} and $\hat{\gamma}=O_{p}(n^{-1/2})$,
one has 
\[
\hat{\gamma}=\frac{\sigma_{n}^{-1}\sum_{i=1}^{n}V_{n}^{(i)}(\theta)}{\sigma_{n}^{-2}\sum_{i=1}^{n}V_{n}^{(i)}(\theta)^{2}}+o_{p}(n^{-1/2}).
\]
Using this expansion for $\hat{\gamma}$, a Taylor expansion argument yields
\begin{align}
2\sum_{i=1}^{n}\log\{1+\hat{\gamma}\sigma_{n}^{-1}V_{n}^{(i)}(\theta)\} & =  2\sum_{i=1}^{n}\left[\hat{\gamma}\sigma_{n}^{-1}V_{n}^{(i)}(\theta)-\frac{1}{2}\{\hat{\gamma}\sigma_{n}^{-1}V_{n}^{(i)}(\theta)\}^{2}\right]+o_{p}(1)\nonumber\\
 & =  \frac{\left\{ \frac{1}{\sigma_{n}\sqrt{n}}\sum_{i=1}^{n}V_{n}^{(i)}(\theta)\right\}^{2}}{\frac{1}{\sigma_{n}^{2}n}\sum_{i=1}^{n}V_{n}^{(i)}(\theta)^{2}}+o_{p}(1).
\label{expansion:V}
\end{align}
Thus, it suffices to show that
\begin{align}
\frac{1}{\sigma_{n}\sqrt{n}}\sum_{i=1}^{n}V_{n}^{(i)}(\theta) & \overset{d}{\to}  \mathcal{N}(0,1),\label{pf:n1-1}\\
\frac{1}{\sigma_{n}^{2}n}\sum_{i=1}^{n}V_{n}^{(i)}(\theta)^{2} & \overset{p}{\to} 1. \label{pf:n1-2}
\end{align}
These two are shown in Lemmas \ref{lem:V} and \ref{lem:VV_standard}, and thus follows the conclusion. \qed \\

\subsection{Proof for Theorem \ref{thm:mJEL} (i)}\label{app:proof:thm:mJEL(i)}
As in the proof of Theorem \ref{thm:JEL}, one can show the asymptotic equivalence in \eqref{expansion:V}. Thus, it suffices to show \eqref{pf:n1-1} and 
\begin{equation} \label{pf:n2-0}
\frac{1}{\sigma_n^2n}\sum_{i=1}^{n}V_{n}^{(i)}(\theta)^{2}\overset{p}{\to} 1+C_*, 
\end{equation}
under both dense- and sparse-subsampling asymptotics $\frac{n}{N_n}\frac{\zeta_{s}}{s\zeta_{1}} \to C_* \in [0, \infty)$. The conclusion then follows from Lemmas \ref{lem:V} and \ref{lem:VV_nonstandard}. \qed \\

\subsection{Proof for Theorem \ref{thm:mJEL} (ii)}\label{app:proof:thm:mJEL(ii)}
As in the proof of Theorem 1, one can show
the asymptotic equivalence
\begin{equation}
\tilde{\ell}(\theta)=\frac{\left\{ \frac{1}{\sigma_{n}\sqrt{n}}\sum_{i=1}^{n}\tilde{V}_{n}^{(i)}(\theta)\right\} ^{2}}{\frac{1}{\sigma_{n}^{2}n}\sum_{i=1}^{n}\tilde{V}_{n}^{(i)}(\theta)^{2}}+o_{p}(1). \label{expansion:V_tilde}
\end{equation}
Thus, it suffices to show that 
\begin{align}
\frac{1}{\sigma_{n}\sqrt{n}}\sum_{i=1}^{n}\tilde{V}_{n}^{(i)}(\theta) & \overset{d}{\to}  \mathcal{N}(0,1+C_*),\label{pf:n2-1}\\
\frac{1}{\sigma_{n}^{2}n}\sum_{i=1}^{n}\tilde{V}_{n}^{(i)}(\theta)^{2} & \overset{p}{\to} 1+C_*. \label{pf:n2-2}
\end{align}
Using the relation $V_n^{(i)}(\hat{\theta})=(n-1)(\hat{\theta}-\hat{\theta}^{(i)})$, the same argument as the proof of Lemma \ref{lem:VV_nonstandard} yields 
\begin{equation} \label{pf:n2-3}
\frac{1}{\sigma_n^2n}\sum_{i=1}^{n}V_{n}^{(i)}(\hat{\theta})^{2}=\frac{1}{\sigma_n^2n}\sum_{i=1}^{n}V_{n}^{(i)}(\theta)^{2}+o_p(1)\overset{p}{\to}1+C_*.
\end{equation}
Thus, the consistency $\hat{\theta}\overset{p}{\to}\theta$ implies (\ref{pf:n2-2}).

It remains to show (\ref{pf:n2-1}). Since $\sum_{i=1}^{n}V_{n}^{(i)}(\hat{\theta})=0$, we have 
\begin{equation*}
    \frac{1}{\sigma_n\sqrt{n}}\sum_{i=1}^{n}\tilde{V}_{n}^{(i)}(\theta) = \sqrt{\frac{\frac{1}{\sigma_n^2}\hat{\mathcal{V}}_{1}}{\frac{1}{\sigma_n^2}(\hat{\mathcal{V}}_{1}-\hat{\mathcal{V}}_{2})}}\frac{\sqrt{n}}{\sigma_n}(\hat{\theta}-\theta).
\end{equation*}
Lemma \ref{lem:CLT} implies $\frac{\sqrt{n}}{\sigma_n}(\hat{\theta}-\theta)\overset{d}{\to}\mathcal{N}(0,1)$. Thus, (\ref{pf:n2-1}) follows from (\ref{pf:n2-3}) and 
\begin{align*}
&\frac{1}{\sigma_n^2}(\hat{\mathcal V}_1-\hat{\mathcal V}_2) = \frac{1}{\sigma_n^2}\left\{\frac{1}{n}\sum_{i=1}^{n}V_{n}^{(i)}(\theta)^{2}-\frac{ns}{N_n}\zeta_s\right\}\{1+o_p(1)\}\\
&=\frac{1}{\sigma_{n}^{2}}\left[\frac{(n-1)^{2}}{n}\sum_{i=1}^{n}\left\{ \frac{1}{{n-1 \choose s}}\sum_{\iota\in I_{n}^{(i)}}\frac{\rho_{\iota}}{p_{n}}h(Z_{\iota},\omega_{\iota})-\frac{1}{{n \choose s}}\sum_{\iota\in I_{n}}\frac{\rho_{\iota}}{p_{n}}h(Z_{\iota},\omega_{\iota})\right\}^{2}-\frac{ns}{N_n}\zeta_s\right]\{1+o_p(1)\}\\
&\overset{p}{\to}1,
\end{align*}
where the first equality follows from Equation (\ref{pf:n2-3}), Lemma \ref{lem:Bernstein}, and the law of large numbers, the second equality follows from the same argument as the proof of Lemma \ref{lem:VV_nonstandard}, and the convergence follows from Lemma \ref{lem:A1_nonstandard}. \qed \\

\subsection{Proof for Lemma \ref{lem:maxV_sufficient}.}\label{app:proof:lem:maxV_sufficient}
Recall that we assume the kernel takes the form of $h(Z_{1},...,Z_{s},\omega_{1...s})=\sum_{i=1}^{s}S_{i}Y_{i}$ with $\sum_{i=1}^{s}S_{i}=1$ and satisfying $\frac{1}{2k-1}\le S_{i}\mathbb{I}\{S_{i}\ne0\}\le\frac{1}{k}$ for a fixed $k$. This implies 
\[
\max_{\iota\in I_{n}}\left|h(Z_{\iota},\omega_{\iota})\right|\lesssim\max_{1\le i\le n}|Y_{i}|.
\]
Recall that for a sequence of real numbers $\{a_{i}\}_{i=1}^{\infty}$, if $n^{-1}\sum_{i}a_{i}$ is convergent as $n\to\infty$, then $\max_{1\le i\le n}|a_{i}|=o(n)$. By the strong law of large numbers, $n^{-1}\sum_{i=1}^{n}|Y_{i}|^{q}\stackrel{a.s.}{\to}\mathbb{E}[|Y_{1}|^{q}]<\infty$, which then implies $\max_{1\le i\le n}|Y_{i}|^{q}=o(n)$ a.s. and hence $\max_{1\le i\le n}|Y_{i}|=o(n^{1/q})$ a.s. Therefore, using the lower bound of $\zeta_{1}$, we conclude that a.s.
\[
\max_{1\le i\le n}|Y_{i}|=o(n^{1/q})=o((n\zeta_{1})^{1/2}).
\]
Furthermore,
\begin{align*}
     \|\E[h(Z_1,...,Z_s,\omega)^2|Z_1]\|_\infty\le \max_{1\le i \le n}|Y_i|^2=\left(\max_{1\le i \le n}|Y_i|^q\right)^{2/q}=o(n^{2/q})\quad a.s.
\end{align*}
where the last equality follows from the arguments above. For \(n^{2/q}\lesssim \frac{n}{s}\), it suffices that $s\lesssim n^{1-2/q}$. \qed \\
\subsection{Proof for Theorem \ref{thm:mJEL-honestRF}.}\label{app:proof:thm:mJEL-honestRF}
We apply Theorem \ref{thm:mJEL} by verifying Assumptions \ref{asm:JEL} and \ref{asm:mJEL}.

Assumption \ref{asm:JEL} (i) is directly implied by the assumption of Theorem \ref{thm:mJEL-honestRF}.

Both Assumptions \ref{asm:JEL} (ii) and \ref{asm:mJEL} follow from Lemma \ref{lem:maxV_sufficient}, as Assumption \ref{asm:lem:maxV_sufficient} required by Lemma \ref{lem:maxV_sufficient} is satisfied. To see this, note that by invoking \citet[Lemma 4]{wager2018estimation}, we have
\[\zeta_1\gtrsim \frac{1}{s(\log s)^d}\gtrsim \frac{1}{n^{1-2/q}},\]
while the rest of Assumption \ref{asm:lem:maxV_sufficient} follows directly under the assumption of Theorem \ref{thm:mJEL-honestRF}.

Assumption \ref{asm:JEL} (iii) can be verified using \citet[Theorem 5]{wager2018estimation} whose assumptions are satisfied by the assumptions of Theorem \ref{thm:mJEL-honestRF}. Under \citet[Theorem 5]{wager2018estimation}, we have
\begin{align}
    \frac{\zeta_s}{s\zeta_1}\lesssim (\log s)^d,\label{eq:WA18-Theorem_5}
\end{align}
from which follows that  \[\frac{s}{n} \frac{\zeta_s}{s\zeta_1}=o(1).\]

Now, we have verified all the requirements for Theorem \ref{thm:mJEL}. By invoking Theorem \ref{thm:mJEL}, we have shown the desired result for the modified EL statistic $\tilde \ell (\theta_n)$. Finally, by the bias bound from \citet[Theorem 3]{wager2018estimation}, one can replace $\theta_n$ in the conclusion by $\theta_0$ (i.e., $\tilde{\ell}(\theta_0)\overset{d}{\to}\chi_{1}^{2}$).
\qed \\
\section{Auxiliary lemmas}

\begin{lemma}
\label{lem:V} Under Assumption \ref{asm:JEL}, it holds that
\[
\frac{1}{\sigma_{n}\sqrt{n}}\sum_{i=1}^{n}V_{n}^{(i)}(\theta)\overset{d}{\to}\mathcal{N}(0,1).
\]
\end{lemma}

\noindent\textbf{Proof.} Observe that 
\begin{equation}
\frac{1}{\sigma_{n}\sqrt{n}}\sum_{i=1}^{n}V_{n}^{(i)}(\theta)  =  \frac{\sqrt{n}}{\sigma_{n}}S_{n}(\theta)-\frac{n-1}{\sigma_{n}\sqrt{n}}\sum_{i=1}^{n}\{S_{n}^{(i)}(\theta)-S_{n}(\theta)\}
  =  \frac{\sqrt{n}}{\sigma_{n}}(\hat{\theta}-\theta),
  \end{equation}
where the second equality follows from
\begin{align}\label{eq:jack}
\sum_{i=1}^n \{S^{(i)}(\theta)-S_n(\theta)\} &=\sum_{i=1}^n\left\{\frac{1}{B_{n}^{(1)}}\sum_{\iota \in I_{1n}^{(i)}} h_n(Z_{\iota},\omega_{\iota})-\hat{\theta}\right\}\nonumber\\
& = \frac{n-s}{B_{n}^{(1)}}\sum_{\iota \in I_{1n}} h_n(Z_{\iota},\omega_{\iota})-n\hat{\theta} =\frac{n}{B_{n}}\sum_{\iota \in I_{1n}} h_n(Z_{\iota},\omega_{\iota})-n\hat{\theta}
=0.
\end{align}
Thus, the conclusion follows from Lemma \ref{lem:CLT}. \qed \\

\begin{lemma}\label{lem:VV_standard}
Suppose that Assumption \ref{asm:JEL} holds true. Then under dense-subsampling asymptotics of 
 $\frac{n}{N_n}\frac{\zeta_{s}}{s\zeta_{1}} =o(1)$, we have
\[
\frac{1}{\sigma_{n}^{2}n}\sum_{i=1}^{n}V_{n}^{(i)}(\theta)^{2}\overset{p}{\to}1.
\]
\end{lemma}

\noindent \textbf{Proof.} Observe that 
\begin{align*}
\frac{1}{\sigma_{n}^{2}n}\sum_{i=1}^{n}V_{n}^{(i)}(\theta)^{2} & =  \frac{1}{\sigma_{n}^{2}n}\sum_{i=1}^{n}\left[(n-1)\{S_{n}(\theta)-S_{n}^{(i)}(\theta)\}+S_{n}(\theta)\right]^{2}\\
 & =  \frac{(n-1)^{2}}{\sigma_{n}^{2}n}\sum_{i=1}^{n}\{S_{n}(\theta)-S_{n}^{(i)}(\theta)\}^{2}+\frac{1}{\sigma_{n}^{2}}S_{n}(\theta)^{2}\\
 & =:  A_{1}+A_{2},
\end{align*}
where the second equality follows from  (\ref{eq:jack}). For $A_{2}$, Lemma \ref{lem:CLT} implies 
\[
A_{2}=\left(\frac{\hat{\theta}-\theta}{\sigma_{n}}\right)^{2}\overset{p}{\to}0.
\]
Thus, it suffices for the conclusion to show that $A_{1}\overset{p}{\to}1$.

Let $N_n^{(1)} = {\mathbb E}[B_{n}^{(1)}]=\frac{n-s}{n}{\mathbb E}[B_n]=\frac{n-s}{n}p_n{n \choose s}=p_n{n-1 \choose s}$. 
Note that 
\begin{align*}
S_{n}^{(i)}(\theta)-S_{n}(\theta)
 & =  \hat{\theta}^{(i)}-\hat{\theta}\\
 & =  \left\{ 1+\left(\frac{N_{n}^{(1)}}{B_{n}^{(1)}}-1\right)\right\} \frac{1}{N_{n}^{(1)}}\sum_{\iota\in I_{n}^{(i)}}\rho_{\iota}h(Z_{\iota},\omega_{\iota})\\
 &\qquad-\left\{ 1+\left(\frac{N_{n}}{B_{n}}-1\right)\right\} \frac{1}{N_{n}}\sum_{\iota\in I_{n}}\rho_{\iota}h(Z_{\iota},\omega_{\iota})\\
 & =  \left\{ 1+\left(\frac{N_{n}}{B_{n}}-1\right)\right\} \left\{ \frac{1}{{n-1 \choose s}}\sum_{\iota\in I_{n}^{(i)}}\frac{\rho_{\iota}}{p_{n}}h(Z_{\iota},\omega_{\iota})-\frac{1}{{n \choose s}}\sum_{\iota\in I_{n}}\frac{\rho_{\iota}}{p_{n}}h(Z_{\iota},\omega_{\iota})\right\},
  \end{align*}
where the third equality follows from the facts that $N_n=p_n{n \choose s}$, $N_n^{(1)}=p_n{n-1 \choose s}$, and $\frac{N_n^{(1)}}{B_{n}^{(1)}}=\frac{N_n}{B_n}$.
Thus, Lemma \ref{lem:Bernstein} implies
\begin{equation*}
A_1=\frac{(n-1)^{2}}{\sigma_{n}^{2}n}\sum_{i=1}^{n}\left\{ \frac{1}{{n-1 \choose s}}\sum_{\iota\in I_{n}^{(i)}}\frac{\rho_{\iota}}{p_{n}}h(Z_{\iota},\omega_{\iota})-\frac{1}{{n \choose s}}\sum_{\iota\in I_{n}}\frac{\rho_{\iota}}{p_{n}}h(Z_{\iota},\omega_{\iota})\right\}^{2}\left\{1+o_{p}(1)\right\},
\end{equation*}
so the conclusion follows from Lemma \ref{lem:A1_standard}. \qed \\

\begin{lemma}\label{lem:VV_nonstandard}
Suppose that Assumptions \ref{asm:JEL} and \ref{asm:mJEL} hold true. Then under both dense-and sparse-subsampling asymptotics,
$\frac{n}{N_n}\frac{\zeta_{s}}{s\zeta_{1}} \to C_* \in [0,\infty)$, we have
\[
\frac{1}{\sigma_n^{2}n}\sum_{i=1}^{n}{V}_{n}^{(i)}(\theta)^{2}\overset{p}{\to} 1+C_*.
\]
\end{lemma}

\noindent \textbf{Proof.} As in the proof of Lemma \ref{lem:VV_standard}, it holds
\begin{equation*}
\frac{1}{\sigma_n^{2}n}\sum_{i=1}^{n}{V}_{n}^{(i)}(\theta)^{2}=A_1+o_p(1).
\end{equation*}
Thus, it suffices to show that $A_1\overset{p}{\to} 1+C_*$. The conclusion then follows from Lemma \ref{lem:A1_nonstandard}. \qed \\

The following result controls the order of the maximal pseudo value under some rate conditions. 
Although the rate conditions are high level, they are rather mild.
As an illustration, suppose the Hajek projection component dominates,
i.e. $\mathbb{V}(U_{n})\asymp s^{2}\zeta_{1}/n$, and suppose that $s\zeta_{1}\asymp\zeta_{s}=O(1)$,
then $\zeta_{1}\asymp1/s$. In this case, if $\mathbb{E}[|Y_{1}|^{q}]\le\infty$,
then 
\[
\zeta_{1}\asymp\frac{1}{s}\gtrsim\frac{C}{n^{1-2/q}}.
\]
For a large $q$, $s$ only needs to be of a slightly smaller order
than $n$.  
\begin{lemma}\label{lem:maxV} 
Under Assumption \ref{asm:JEL}, it holds 
\[
\max_{1\le i\le n}|\sigma_{n}^{-1}V_{n}^{(i)}(\theta)|=o_{p}(n^{1/2}).
\]
\end{lemma}

\noindent \textbf{Proof.} We normalize $\theta=0$ without loss of generality. Recall that 
\[
V_{n}^{(i)}(\theta)=\frac{n}{B_{n}}\sum_{\iota\in I_{n}}\rho_{\iota}h(Z_{\iota},\omega_{\iota})-(n-1)\frac{1}{B_n^{(1)}}\sum_{\iota\in I_{n}^{(i)}}\rho_{\iota}h(Z_{\iota},\omega_{\iota}).
\]
By Lemma \ref{lem:Bernstein}, it suffices to show the result with
$B_{n}$, $B_n^{(1)}$ replaced by $N_{n}$, $N_{n}^{(1)}$ for
all $i=1,...,n$, respectively.

For notational convenience, define $h(Z_{i_{1}...i_{s}},\omega_{1...s})$ and $\rho_{i_{1}...i_{s}}$ for those vectors of indices $(i_{1},...,i_{s})$ that do not satisfy $i_{1}<...<i_{s}$ to be equal to $h(Z_{\iota},\omega_{\iota})$ and $\rho_{\iota}$, respectively, where $\iota$ consists of increasingly sorted $\{i_{1},...,i_{s}\}$. Also, define the shorthand notation of $h_{\iota}=h(Z_{\iota},\omega_{\iota})$. Following some rearrangement,
\begin{eqnarray*}
\frac{1}{N_{n}}\sum_{\iota\in I_{n}}\rho_{\iota}h_{\iota} & = & \frac{1}{p_{n}n...(n-s+1)}\sum_{\iota:\:\text{distinctive }i_{j}}\rho_{\iota}h_{\iota}\\
 & = & \frac{s}{p_{n}n...(n-s+1)}\sum_{\ell\in J_{n}^{(i)}}\rho_{(i,\ell)}h_{(i,\ell)}+\frac{{n-1 \choose s}}{{n \choose s}}\frac{1}{p_{n}{n-1 \choose s}}\sum_{\iota\in I_{n}^{(i)}}\rho_{\iota}h_{\iota},
\end{eqnarray*}
where the index $(i,\ell)=(i,\ell_{1},...,\ell_{s-1})$ and 
$$J_{n}^{(i)}=\{(\ell_{1},...,\ell_{s-1}):1\le\ell_{1},...,\ell_{s-1}\le n\text{ all distinctive and \ensuremath{\ne i}}\}.$$
Thus, uniformly over $i=1,...,n$, we have 
\begin{eqnarray*}
|V_{n}^{(i)}(\theta)| & = & \left|\frac{ns}{p_{n}n...(n-s+1)}\sum_{\ell\in J_{n}^{(i)}}\rho_{(i,\ell)}h_{(i,\ell)}-\left((n-1)-\frac{n{n-1 \choose s}}{{n \choose s}}\right)\frac{1}{p_{n}{n-1 \choose s}}\sum_{\iota\in I_{n}^{(i)}}\rho_{\iota}h_{\iota}\right|\\
 & = & \left|\frac{s}{p_{n}(n-1)...(n-s+1)}\sum_{\ell\in J_{n}^{(i)}}\rho_{(i,\ell)}h_{(i,\ell)}-\frac{(s-1)}{p_{n}{n \choose s-1}}\sum_{\iota\in I_{n}^{(i)}}\rho_{\iota}h_{\iota}\right|\\
 & \le & s\max_{\iota\in I_{n}}\left|h(Z_{\iota},\omega_{\iota})\right|\left(\frac{1}{p_{n}(n-1)...(n-s+1)}\sum_{\ell\in J_{n}^{(i)}}\left|\rho_{(i,\ell)}\right|+\frac{1}{p_{n}{n \choose s-1}}\sum_{\iota\in I_{n}^{(i)}}\left|\rho_{\iota}\right|\right)\\
 & \le & s\max_{\iota\in I_{n}}\left|h(Z_{\iota},\omega_{\iota})\right|\cdot O_p(1),
\end{eqnarray*}
where the last inequality follows from  (\ref{eq:bernstein}) and the fact that $\rho_{\iota}$'s are i.i.d. Thus we have, with probability approaching one 
\begin{align*}
\max_{1\le i\le n}|V_{n}^{(i)}(\theta)|\lesssim & s\max_{\iota\in I_{n}}\left|h(Z_{\iota},\omega_{\iota})\right|.
\end{align*}
Hence by Assumption \ref{asm:JEL} and recall that $\sigma_{n}=\sqrt{s^{2}\zeta_{1}}$, we have
\[
\max_{1\le i\le n}|\sigma_{n}^{-1}V_{n}^{(i)}(\theta)|\lesssim \frac{s\max_{\iota\in I_{n}}\left|h(Z_{\iota},\omega_{\iota})\right|}{\sqrt{s^2\zeta_{1}
}}=o_p(n^{1/2}).
\]
\qed \\

\begin{lemma} \label{lem:A1_standard} 
Suppose that Assumption \ref{asm:JEL} holds true. Then under dense-subsampling asymptotics, $\frac{n}{N_n}\frac{\zeta_{s}}{s\zeta_{1}} =o(1)$, we have
\[
\frac{(n-1)^{2}}{\sigma_{n}^{2}n}\sum_{i=1}^{n}\left\{ \frac{1}{{n-1 \choose s}}\sum_{\iota\in I_{n}^{(i)}}\frac{\rho_{\iota}}{p_{n}}h_n(Z_{\iota},\omega_{\iota})-\frac{1}{{n \choose s}}\sum_{\iota\in I_{n}}\frac{\rho_{\iota}}{p_{n}}h_n(Z_{\iota},\omega_{\iota})\right\}^{2}\stackrel{p}{\to}1.
\]
\end{lemma}

\noindent \textbf{Proof.} 
Note that for each $j=1,...,s$, 
\begin{align}
 &   {n \choose j}^{-1}\sum_{(n,j)}h^{(j)}(Z_{i_{1}},\ldots,Z_{i_{j}})-{n-1 \choose j}^{-1}\sum_{(n-1,j)}^{(i)}h^{(j)}(Z_{i_{1}},\ldots,Z_{i_{j}})\nonumber\\
 & =  {n \choose j}^{-1}\sum_{i_{1}<\ldots<i_{j}:\:\exists i}h^{(j)}(Z_{i_{1}},\ldots,Z_{i_{j}})+\left\{ {n \choose j}^{-1}-{n-1 \choose j}^{-1}\right\} \sum_{(n-1,j)}^{(i)}h^{(j)}(Z_{i_{1}},\ldots,Z_{i_{j}})\nonumber\\
 & =  {n \choose j}^{-1}\sum_{i_{1}<\ldots<i_{j}:\:\exists i}h^{(j)}(Z_{i_{1}},\ldots,Z_{i_{j}})-\frac{j}{n-j}{n \choose j}^{-1}\sum_{(n-1,j)}^{(i)}h^{(j)}(Z_{i_{1}},\ldots,Z_{i_{j}}), \label{eq:U_LO}
\end{align}
and
\begin{align}
\tilde{U}_n-\tilde{U}_n^{(i)}
 &     =  {n \choose s}^{-1}\sum_{i_{1}<\ldots<i_{s}:\:\exists i}\left(\frac{\rho_{\iota}}{p_n}-1\right)h(Z_\iota,\omega_{\iota})\nonumber\\
    &\quad+\left\{ {n \choose s}^{-1}-{n-1 \choose s}^{-1}\right\} \sum_{(n-1,s)}^{(i)}\left(\frac{\rho_{\iota}}{p_n}-1\right)h(Z_\iota,\omega_{\iota})\nonumber\\
     &  =  {n \choose s}^{-1}\sum_{i_{1}<\ldots<i_{s}:\:\exists i}\left(\frac{\rho_{\iota}}{p_n}-1\right)h(Z_{i_{1}},\ldots,Z_{i_{s}},\omega_{\iota})\nonumber\\
     &\quad-\frac{s}{n-s}{n \choose s}^{-1}\sum_{(n-1,s)}^{(i)}\left(\frac{\rho_{\iota}}{p_n}-1\right)h(Z_{i_{1}},\ldots,Z_{i_{s}},\omega_{\iota}). \label{eq:U_tilde_LO}
\end{align}
By using Equations (\ref{eq:Hoeffding}), (\ref{eq:U_LO}), and (\ref{eq:U_tilde_LO}), we decompose
\begin{align*}
 &\frac{(n-1)^{2}}{\sigma_{n}^{2}n}\sum_{i=1}^{n}\left\{ \frac{1}{{n \choose s}}\sum_{\iota\in I_{n}}\frac{\rho_{\iota}}{p_{n}}h(Z_{\iota},\omega_{\iota})-\frac{1}{{n-1 \choose s}}\sum_{\iota\in I_{n}^{(i)}}\frac{\rho_{\iota}}{p_{n}}h(Z_{\iota},\omega_{\iota})\right\} ^{2}\\
 & =  \frac{(n-1)^{2}}{\sigma_{n}^{2}n}\sum_{i=1}^{n}(U_n-U_n^{(i)}+\tilde{U}_n-\tilde{U}_n^{(i)})^2\\
 &=: \sum_{i=1}^n(X_i+Y_i)^2,
\end{align*}
where
\begin{align*}
X_i &=\frac{n-1}{\sigma_n \sqrt{n}}\frac{s}{n}h^{(1)}(Z_i),\\
Y_i&=\frac{n-1}{\sigma_n \sqrt{n}}\left[\begin{array}{c}
\sum_{j=2}^s{s \choose j}\left\{{n \choose j}^{-1}\sum_{i_{1}<\ldots<i_{j}:\:\exists i}h^{(j)}(Z_{i_{1}},\ldots,Z_{i_{j}})\right\}\\
-\sum_{j=1}^s{s \choose j}\left\{\frac{j}{n-j}{n \choose j}^{-1}\sum_{(n-1,j)}^{(i)}h^{(j)}(Z_{i_{1}},\ldots,Z_{i_{j}})\right\}\\
+{n \choose s}^{-1}\sum_{i_{1}<\ldots<i_{s}:\:\exists i}\left(\frac{\rho_{\iota}}{p_n}-1\right)h(Z_{i_{1}},\ldots,Z_{i_{s}},\omega_{\iota})\\
-\frac{s}{n-s}{n \choose s}^{-1}\sum_{(n-1,s)}^{(i)}\left(\frac{\rho_{\iota}}{p_n}-1\right)h(Z_{i_{1}},\ldots,Z_{i_{s}},\omega_{\iota})
\end{array}\right].
 \end{align*}
Thus, the conclusion follows by applying Lemma \ref{lem:stefanski_1}.

It remains to verify the conditions for Lemma \ref{lem:stefanski_1}. Since $\sum_{i=1}^n {\mathbb E}[X_i^2]\to 1$, we obtain $\sum_{i=1}^n X_i^2\overset{p}{\to} 1$ by the definition $\sigma_n=\sqrt{s^2\zeta_{1}}$ and the law of large numbers.

We now verify $\sum_{i=1}^n {\mathbb E}[Y_i^2] \to 0$. Using the properties of Hoeffding decompositions, the facts that $\rho_{\iota}$'s are independent of ${\mathcal Z}_n$ and $\omega_{\iota}$, $j \le s$, and $s/n=o(1)$, direct calculation yields
\begin{align}
\sum_{i=1}^n{\mathbb E}[Y_i^2]&=\frac{n}{\sigma_n^2}\sum_{i=1}^n{\mathbb E}\left[\begin{array}{c}
\left\{\sum_{j=2}^s{s \choose j}{n \choose j}^{-1}\sum_{i_{1}<\ldots<i_{j}:\:\exists i}h^{(j)}(Z_{i_{1}},\ldots,Z_{i_{j}})\right\}^2\nonumber\\
+\left\{{n \choose s}^{-1}\sum_{i_{1}<\ldots<i_{s}:\:\exists i}\left(\frac{\rho_{\iota}}{p_n}-1\right)h(Z_{i_{1}},\ldots,Z_{i_{s}},\omega_{\iota})\right\}^2
\end{array}\right]\{1+o(1)\}\nonumber\\
&=\frac{n^2}{\sigma_n^2}\left[\sum_{j=2}^s {s \choose j}^2{n\choose j}^{-2}{n-1\choose j-1}V_j
+{n \choose s}^{-2}\frac{1-p_n}{p_n}{n-1\choose s-1}\zeta_{s}\right]\{1+o(1)\}\nonumber\\
&=\frac{n^2}{\sigma_n^2}\left[\sum_{j=2}^s \frac{j}{n} {s \choose j}^2{n\choose j}^{-1}V_j
+\frac{s}{nN_n}(1-p_n)\zeta_{s}\right]\{1+o(1)\}, \label{eq:Y}
\end{align}
where the second equality uses the fact that the set $\{i_{1}<\ldots<i_{s}:\exists i\}$ has the cardinality of ${n-1 \choose s-1}$ and the third equality follows from the facts that ${n-1 \choose s-1}=\frac{s}{n}{n\choose s}$ and $N_n=p_n{n \choose s}$. 
Note that for $n$ large enough, the first term of the right hand side in (\ref{eq:Y}) is written as
\begin{align}
\frac{n}{s^2\zeta_{1}}\sum_{j=2}^{s}j{s \choose j}^{2}{n \choose j}^{-1}V_{j}
 & =  \frac{1}{\zeta_{1}}\sum_{j=2}^{s}\frac{j}{s}{s-1 \choose j-1}{n-1 \choose j-1}^{-1}{s \choose j}V_{j}\nonumber \\
 & \le  \frac{1}{\zeta_{1}}\sum_{j=2}^{s}\frac{2}{n}{s \choose j}V_{j} = \frac{2}{n\zeta_{1}}(\zeta_{s}-s\zeta_{1}), \label{eq:Efron-Stein}
\end{align}
where the last equality follows from the identity $\zeta_{s}=\sum_{j=1}^{s}{s \choose j}V_{j}$. To see why the inequality holds, note that for $j=2$, 
\[
\frac{2}{s}{s-1 \choose 1}{n-1 \choose 1}^{-1}=\frac{2}{s}\cdot\frac{(s-1)}{(n-1)}\le\frac{2}{n},
\]
and for all $j=3,...,s$, 
\begin{align*}
   \frac{j}{s}{s-1 \choose j-1}{n-1 \choose j-1}^{-1}=&\frac{j}{s}\cdot\frac{(s-1)\cdots(s-j+1)}{(n-1)\cdots(n-j+1)}\\
   \le&\frac{j}{n-1}\cdot\frac{(s-2)\cdots(s-j+1)}{(n-2)\cdots(n-j+1)}\\
   \le&\frac{j}{n-1}\left(\frac{1}{2}\right)^{j-2}\le\frac{2}{n}, 
\end{align*}
as $s\le n/2$ for $n$ large enough. Now, under our assumption,
\eqref{eq:Efron-Stein} satisfies 
\[
\frac{1}{n\zeta_{1}}(\zeta_{s}-s\zeta_{1})=\frac{s}{n}\left(\frac{\zeta_{s}}{s\zeta_{1}}-1\right)=o(1).
\]
For the second term of (\ref{eq:Y}), we have
\begin{equation*}
\frac{ns}{\sigma_n^2 N_n}(1-p_n)\zeta_{s}=\frac{n}{N_n}\frac{\zeta_{s}}{s\zeta_{1}}(1-p_n) =o(1).
\end{equation*}
under the condition $\frac{n}{N_n}\frac{\zeta_{s}}{s\zeta_{1}} =o(1)$. Thus, the conclusion follows by Lemma \ref{lem:stefanski_1}. \qed \\

\begin{lemma} \label{lem:A1_nonstandard}
Suppose Assumptions \ref{asm:JEL} and \ref{asm:mJEL} are satisfied. Then under both dense- and sparse-subsampling asymptotics,
$\frac{n}{N_n}\frac{\zeta_{s}}{s\zeta_{1}} \to C_* \in [0,\infty)$,
we have
\[
\frac{(n-1)^{2}}{\left(\sigma_{n}^{2}+\frac{ns\zeta_{s}}{N_n}\right)n}\sum_{i=1}^{n}\left\{ \frac{1}{{n-1 \choose s}}\sum_{\iota\in I_{n}^{(i)}}\frac{\rho_{\iota}}{p_{n}}h_n(Z_{\iota},\omega_{\iota})-\frac{1}{{n \choose s}}\sum_{\iota\in I_{n}}\frac{\rho_{\iota}}{p_{n}}h_n(Z_{\iota},\omega_{\iota})\right\}^{2}\stackrel{p}{\to}1.
\]
\end{lemma}

\noindent \textbf{Proof.} We will apply Lemma \ref{lem:stefanski_1} by setting 
\begin{align*}
\tilde{X}_i &=\frac{n-1}{\sqrt{\left(\sigma_{n}^{2}+\frac{ns\zeta_{s}}{N_n}\right)n}}\left\{\frac{s}{n}h^{(1)}(Z_i)+{n \choose s}^{-1}\sum_{i_{1}<\ldots<i_{s}:\:\exists i}\left(\frac{\rho_{\iota}}{p_n}-1\right)h(Z_{i_{1}},\ldots,Z_{i_{s}},\omega_\iota)\right\},\\
\tilde{Y}_i&=\frac{n-1}{\sqrt{\left(\sigma_{n}^{2}+\frac{ns\zeta_{s}}{N_n}\right)n}}\left[\begin{array}{c} \sum_{j=2}^s{s \choose j}\left\{{n \choose j}^{-1}\sum_{i_{1}<\ldots<i_{j}:\:\exists i}h^{(j)}(Z_{i_{1}},\ldots,Z_{i_{j}})\right\}\\
-\sum_{j=1}^s{s \choose j}\left\{\frac{j}{n-j}{n \choose j}^{-1}\sum_{(n-1,j)}^{(i)}h^{(j)}(Z_{i_{1}},\ldots,Z_{i_{j}})\right\}\\
-\frac{s}{n-s}{n \choose s}^{-1}\sum_{(n-1,s)}^{(i)}\left(\frac{\rho_{\iota}}{p_n}-1\right)h(Z_{i_{1}},\ldots,Z_{i_{s}},\omega_\iota)
\end{array}\right].
 \end{align*}
 Since $\sum_{i=1}^n {\mathbb E}[\tilde{Y}_i^2] \to 0$ follows from the proof of (i), it suffices to show that $\sum_{i=1}^n {\mathbb E}[\tilde{X}_i^2]\to 1$ and $\sum_{i=1}^n \tilde{X}_i^2\overset{p}{\to} 1$.
Note that
\begin{align*}
\sum_{i=1}^n {\mathbb E}[\tilde{X}_i^2]
&=\frac{(n-1)^2}{s^2\zeta_{1}+\frac{ns\zeta_{s}}{N_n}}\left\{\frac{s^2}{n^2}\zeta_{1}+\frac{s}{nN_n}(1-p_n)\zeta_{s}\right\}
\to 1.
\end{align*}
and 
\begin{align*}
&\sum_{i=1}^n \tilde{X}_i^2-1\\
=& \frac{(n-1)^2}{\left(s^2\zeta_{1}+\frac{ns\zeta_{s}}{N_n}\right)n}\sum_{i=1}^n\left\{\frac{s}{n}h^{(1)}(Z_i)+{n \choose s}^{-1}\sum_{i_{1}<\ldots<i_{s}:\:\exists i}\left(\frac{\rho_{\iota}}{p_n}-1\right)h(Z_{i_{1}},\ldots,Z_{i_{s}},\omega_\iota)\right\}^2-1\\
\le & \frac{n}{s^2\zeta_{1}+\frac{ns\zeta_{s}}{N_n}}\sum_{i=1}^n\left\{\frac{s}{n}h^{(1)}(Z_i)+{n \choose s}^{-1}\sum_{i_{1}<\ldots<i_{s}:\:\exists i}\left(\frac{\rho_{\iota}}{p_n}-1\right)h(Z_{i_{1}},\ldots,Z_{i_{s}},\omega_\iota)\right\}^2-1\\
=& \frac{1}{\frac{s^2\zeta_{1}}{n}+\frac{s\zeta_{s}}{N_n}}\frac{1}{n}\sum_{i=1}^n\frac{s^2}{n}(h^{(1)}(Z_i)^2-\zeta_{1})\\
&+\frac{1}{\frac{s^2\zeta_{1}}{n}+\frac{s\zeta_{s}}{N_n}}\sum_{i=1}^n\left[\left\{{n \choose s}^{-1}\sum_{i_{1}<\ldots<i_{s}:\:\exists i}\left(\frac{\rho_{\iota}}{p_n}-1\right)h(Z_{i_{1}},\ldots,Z_{i_{s}},\omega_\iota)\right\}^2-\frac{s\zeta_{s}}{nN_n}\right]\\
&+\frac{2}{\frac{s^2\zeta_{1}}{n}+\frac{s\zeta_{s}}{N_n}}\sum_{i=1}^n\left\{\frac{s}{n}h^{(1)}(Z_i)\right\}\left\{{n \choose s}^{-1}\sum_{i_{1}<\ldots<i_{s}:\:\exists i}\left(\frac{\rho_{\iota}}{p_n}-1\right)h(Z_{i_{1}},\ldots,Z_{i_{s}},\omega_\iota)\right\}\\
&=:(i).
\end{align*}
To evaluate the components of $(i)$, note that for its first term, \[\frac{1}{\frac{s^2\zeta_1}{n}+\frac{s\zeta_s}{N_n}}\frac{1}{n}\sum_{i=1}^n\frac{s^2}{n}(h^{(1)}(Z_i)^2-\zeta_1)=o_p(1)\]
by the law of large numbers.
{
For the second term in $(i)$, 
\begin{align*}
&{\mathbb E}\left[\left(\frac{1}{\frac{s^2\zeta_1}{n}+\frac{s\zeta_s}{N_n}}\sum_{i=1}^n\left[\left\{{n \choose s}^{-1}\sum_{i_{1}<\ldots<i_{s}:\:\exists i}\left(\frac{\rho_{\iota}}{p_n}-1\right)h(Z_{i_{1}},\ldots,Z_{i_{s}},\omega_\iota)\right\}^2-\frac{s\zeta_s}{nN_n}\right]\right)^2\right]\\
=& \frac{1}{\left(\frac{s^2\zeta_1}{n}+\frac{s\zeta_s}{N_n}\right)^2}{\mathbb E}\left[\left(\sum_{i=1}^n\left\{{n \choose s}^{-1}\sum_{i_{1}<\ldots<i_{s}:\:\exists i}\left(\frac{\rho_{\iota}}{p_n}-1\right)h(Z_{i_{1}},\ldots,Z_{i_{s}},\omega_\iota)\right\}^2-\frac{s\zeta_s}{N_n}\right)^2\right]\\
= &\frac{1}{\left(\frac{s^2\zeta_1}{n}+\frac{s\zeta_s}{N_n}\right)^2}\left({\mathbb E}\left[\left(\sum_{i=1}^n\left\{{n \choose s}^{-1}\sum_{i_{1}<\ldots<i_{s}:\:\exists i}\left(\frac{\rho_{\iota}}{p_n}-1\right)h(Z_{i_{1}},\ldots,Z_{i_{s}},\omega_\iota)\right\}^2\right)^2\right]+\{1-2(1-p_n)\}\frac{s^2\zeta_s^2}{N_n^2}\right)\\
= &\frac{1}{\left(\frac{s^2\zeta_1}{n}+\frac{s\zeta_s}{N_n}\right)^2}\left(\sum_{i=1}^n\sum_{j\neq i}^n{\mathbb E}\left[\begin{array}{c}\left\{{n \choose s}^{-1}\sum_{i_{1}<\ldots<i_{s}:\:\exists i}\left(\frac{\rho_{\iota}}{p_n}-1\right)h(Z_{i_{1}},\ldots,Z_{i_{s}},\omega_\iota)\right\}^2\\ \times\left\{{n \choose s}^{-1}\sum_{i_{1}'<\ldots<i_{s}':\:\exists j}\left(\frac{\rho_{\iota'}}{p_n}-1\right)h(Z_{i_{1}'},\ldots,Z_{i_{s}'},\omega_{\iota'})\right\}^2\end{array}\right]+(2p_n-1)\frac{s^2\zeta_s^2}{N_n^2}\right)\\
&+\frac{1}{\left(\frac{s^2\zeta_1}{n}+\frac{s\zeta_s}{N_n}\right)^2}\left(\sum_{i=1}^n{\mathbb E}\left[\left\{{n \choose s}^{-1}\sum_{i_{1}<\ldots<i_{s}:\:\exists i}\left(\frac{\rho_{\iota}}{p_n}-1\right)h(Z_{i_{1}},\ldots,Z_{i_{s}},\omega_\iota)\right\}^4\right]\right)\\
=&\frac{1}{\left(\frac{s^2\zeta_1}{n}+\frac{s\zeta_s}{N_n}\right)^2}\left[n(n-1){n \choose s}^{-4}\left\{{n-1 \choose s-1}^{2}+2{n-2 \choose s-2}^2\right\}\left(\frac{1-p_n}{p_n}\right)^2\zeta_s^2+(2p_n-1)\frac{s^2\zeta_s^2}{N_n^2}\right]\\
&+\frac{1}{\left(\frac{s^2\zeta_1}{n}+\frac{s\zeta_s}{N_n}\right)^2}\left\{n {n \choose s}^{-4}{n-1 \choose s-1}\left(\frac{1-p_n}{p_n^3}-\frac{3(1-p_n)^2}{p_n^2}\right){\mathbb E}[h(Z_1, \ldots, Z_s,\omega)^4]\right\}\\
=&\frac{1}{\left(\frac{s^2\zeta_1}{n}+\frac{s\zeta_s}{N_n}\right)^2}\left\{\frac{n-1}{n}\left(1-p_n\right)^2\frac{s^2\zeta_s^2}{N_n^2}+\frac{2(s-1)^2}{n(n-1)}\left(1-p_n\right)^2\frac{s^2\zeta_s^2}{N_n^2}+(2p_n-1)\frac{s^2\zeta_s^2}{N_n^2}\right\}\{1+o(1)\}\\
&+\frac{1}{\left(\frac{s^2\zeta_1}{n}+\frac{s\zeta_s}{N_n}\right)^2}\left\{s\left(\frac{1-p_n}{N_n^3}-{n \choose s}^{-1}\frac{3(1-p_n)^2}{N_n^2}\right){\mathbb E}[h(Z_1, \ldots, Z_s,\omega)^4]\right\}\\
=& o(1),
\end{align*}
where the fourth equality follows from the fact that $\rho_{\iota}$'s are i.i.d. and independent of ${\mathcal Z}_n$ and $\omega_{\iota}$, ${\mathbb E}[(\rho_{\iota}-p_n)^2]=p_n(1-p_n)$ and ${\mathbb E}[(\rho_{\iota}-p_n)^4]=p_n(1-p_n)\{1-3p_n(1-p_n)\}$, the fifth equality follows from $N_n=p_n{n \choose s}$, ${n-1 \choose s-1}=\frac{s}{n}{n\choose s}$ and ${n-2 \choose s-2}=\frac{s(s-1)}{n(n-1)}{n\choose s}$, and the last equality follows from 
\[\frac{n-1}{n}\left(1-p_n\right)^2\frac{s^2\zeta_s^2}{N_n^2}+(2p_n-1)\frac{s^2\zeta_s^2}{N_n^2}=\left\{p_n^2-\frac{(1-p_n)^2}{n}\right\}\frac{s^2\zeta_s^2}{N_n^2},\] $s/n=o(1)$ and $p_n=o(1)$.


For the third term in $(i)$,
\begin{align*}
&{\mathbb E}\left[\left(\frac{2}{\left(\frac{s^2\zeta_1}{n}+\frac{s\zeta_s}{N_n}\right)}\sum_{i=1}^n\left\{\frac{s}{n}h^{(1)}(Z_i)\right\}\left\{{n \choose s}^{-1}\sum_{i_{1}<\ldots<i_{s}:\:\exists i}\left(\frac{\rho_{\iota}}{p_n}-1\right)h(Z_{i_{1}},\ldots,Z_{i_{s}},\omega_\iota)\right\}\right)^2\right]\\
=& 
\frac{4(\frac{s}{n})^{2}}{\left(\frac{s^2\zeta_1}{n}+\frac{s\zeta_s}{N_n}\right)^2 }\E\left[\left\{ \sum_{i=1}^{n}{n \choose s}^{-1}\sum_{i_{1}<\ldots<i_{s}:\:\exists i}\left(\frac{\rho_{\iota}}{p_{n}}-1\right)h_{1}(Z_{i})h(Z_{i},Z_{i_{1}},\ldots,Z_{i_{s-1}},\omega_\iota)\right\}^{2}\right]\\
=& 
\frac{4(\frac{s}{n})^{2}}{\left(\frac{s^2\zeta_1}{n}+\frac{s\zeta_s}{N_n}\right)^2 }{n \choose s}^{-2}\sum_{i=1}^{n}\sum_{j\neq i}^n\E\left[\begin{array}{c}
\left\{\sum_{i_{1}<\ldots<i_{s}:\:\exists i}\left(\frac{\rho_{\iota}}{p_{n}}-1\right)h_{1}(Z_{i})h(Z_{i},Z_{i_{1}},\ldots,Z_{i_{s-1}},\omega_\iota)\right\} \\
\times\left\{\sum_{i_{1}'<\ldots<i_{s}':\:\exists j}\left(\frac{\rho_{\iota'}}{p_{n}}-1\right)h_{1}(Z_{j})h(Z_{j},Z_{i_{1}'},\ldots,Z_{i_{s-1}'},\omega_{\iota'})\right\} 
\end{array}\right]\\
& +\frac{4(\frac{s}{n})^{2}}{\left(\frac{s^2\zeta_1}{n}+\frac{s\zeta_s}{N_n}\right)^2 }{n \choose s}^{-2}\sum_{i=1}^{n}\E\left[
\left\{\sum_{i_{1}<\ldots<i_{s}:\:\exists i}\left(\frac{\rho_{\iota}}{p_{n}}-1\right)h_{1}(Z_{i})h(Z_{i},Z_{i_{1}},\ldots,Z_{i_{s-1}},\omega_\iota)\right\}^2\right]\\
= &  \frac{4(\frac{s}{n})^{2}}{\left(\frac{s^2\zeta_1}{n}+\frac{s\zeta_s}{N_n}\right)^2 }{n \choose s}^{-2}{n-2 \choose s-2}\left(\frac{1-p_{n}}{p_{n}}\right)n(n-1)\E\left[h_{1}(Z_{1})h_{1}(Z_{2})h(Z_{1},Z_{2}, \ldots, Z_{s},\omega)^2\right]\\
&+ \frac{4(\frac{s}{n})^{2}}{\left(\frac{s^2\zeta_1}{n}+\frac{s\zeta_s}{N_n}\right)^2 }{n \choose s}^{-2}{n-1 \choose s-1}\left(\frac{1-p_{n}}{p_{n}}\right)n\E\left[
h_{1}(Z_{1})^2h(Z_{1},,\ldots, Z_{s},\omega)^2\right]\\
=& \frac{4s^3(s-1)}{n^2N_n\left(\frac{s^2\zeta_1}{n}+\frac{s\zeta_s}{N_n}\right)^2 }\left(1-p_{n}\right)\E\left[h_{1}(Z_{1})h_{1}(Z_{2})h(Z_{1},Z_{2}, \ldots, Z_{s},\omega)^2\right]\\
&+\frac{4s^3}{n^2N_n\left(\frac{s^2\zeta_1}{n}+\frac{s\zeta_s}{N_n}\right)^2 }\left(1-p_{n}\right)\E\left[h_{1}(Z_{1})^2h(Z_{1},\ldots, Z_{s},\omega)^2\right]\\
=:&(ii),
\end{align*}
where the third equality follows from the fact that $\rho_{\iota}$'s are i.i.d. and independent of ${\mathcal Z}_n$ and $\omega_{\iota}$, and the fourth inequality follows form $N_n=p_n{n \choose s}$, ${n\choose s}^{-1}{n-2 \choose s-2}=\frac{s(s-1)}{n(n-1)}$ and ${n\choose s}^{-1}{n-1 \choose s-1}=\frac{s}{n}$. To further simplify the right hand side, notice that  by Cauchy-Schwarz, the symmetry of $h$ and i.i.d. sampling, we have
\[
   \left| \E\left[h_{1}(Z_{1})h_{1}(Z_{2})h(Z_{1},Z_{2}, \ldots, Z_{s},\omega)^2\right]\right|\le 
   \E\left[h_{1}(Z_{1})^2h(Z_{1}, \ldots, Z_{s},\omega)^2\right].
\]
Hence 
\begin{align*}
   (ii) \lesssim& \frac{s^4}{n^2N_n\left(\frac{s^2\zeta_1}{n}+\frac{s\zeta_s}{N_n}\right)^2 }\E\left[h_{1}(Z_{1})^2h(Z_{1},\ldots, Z_{s},\omega)^2\right]\\
    \lesssim&  \frac{1}{N_n\zeta_1^2 }\E\left[h_{1}(Z_{1})^2h(Z_{1},\ldots, Z_{s},\omega)^2\right]\\
    \lesssim& 
    \frac{s}{n\zeta_1\zeta_s } \E\left[h_{1}(Z_{1})^2h(Z_{1},\ldots, Z_{s},\omega)^2\right]\\
     =& 
    \frac{s}{n\zeta_1\zeta_s } \E\left[h_{1}(Z_{1})^2\E[h(Z_{1},\ldots, Z_{s},\omega)^2|Z_1]\right]\\
    \le&
    \frac{s}{n\zeta_1\zeta_s}\zeta_1 \|\E[h(Z_{1},\ldots, Z_{s},\omega)^2|Z_1]\|_\infty\\
    =&    \frac{s}{n\zeta_s} \|\E[h(Z_{1},\ldots, Z_{s},\omega)^2|Z_1]\|_\infty
    =o\left(1\right),
\end{align*}
where the third inequality follows from that  $N_n\gtrsim \frac{n\zeta_s}{s\zeta_1}$ implied by the current regime, the fourth by H\"older's inequality, and the last by $\zeta_s\ge L>0$ and a.s.
\(
   \|\E[h(Z_{1},\ldots, Z_{s},\omega)^2|Z_1]\|_\infty=o\left(\frac{n}{s}\right). 
\)

By combining these results, we now have $(i)=o_p(1)$. Thus, the conclusion follows. \qed \\

}

\begin{lemma}[Bernstein's inequality]
\label{lem:Bernstein} For  any $t>0$, we have
\begin{align}
 & \mathbb{P}\left(\left|\frac{B_{n}}{N_{n}}-1\right|>\sqrt{\frac{2t}{N_{n}}}+\frac{2t}{3N_{n}}\right)\le2e^{-t}.\label{eq:bernstein}
\end{align}
\end{lemma}

\noindent \textbf{Proof.} This is a direct implication of Bernstein's inequality for i.i.d. Bernoulli random variables in \citet[Lemma 2.2.9]{vdVW1996}.
\qed

The following lemma, taken from \citet*[Lemma 1]{peng2021bias}, is useful for establishing results on variance estimation.
\begin{lemma}[Convergence of second moment]
\label{lem:stefanski_1} Suppose that $\sum_{i=1}^{n}X_{i}^{2}\stackrel{p}{\to}1$,
$\sum_{i=1}^{n}\mathbb{E}[X_{i}^{2}]\to1$, and $\sum_{i=1}^{n}\mathbb{E}[Y_{i}^{2}]\to0$.
Then it holds that 
\begin{align*}
    \sum_{i=1}^{n}(X_{i}+Y_{i})^{2}\stackrel{p}{\to}1
\quad\text{ and }\quad\sum_{i=1}^{n}\mathbb{E}[(X_{i}+Y_{i})^{2}]\to 1.
\end{align*}
 
\end{lemma}

\noindent \textbf{Proof.} Although the proof is presented in \citet*{peng2021bias}, it is included for completeness. First observe that 
\[
\sum_{i=1}^{n}(X_{i}+Y_{i})^{2}=\sum_{i=1}^{n}X_{i}^{2}+\sum_{i=1}^{n}Y_{i}^{2}+2\sum_{i=1}^{n}X_{i}Y_{i}.
\]
By Markov's inequality, $\sum_{i=1}^{n}\mathbb{E}[Y_{i}^{2}]\to0$
implies $\sum_{i=1}^{n}Y_{i}^{2}\stackrel{p}{\to}0$. Moreover, by
Cauchy-Schwarz, we have 
\[
\mathbb{E}\left[\left|\sum_{i=1}^{n}X_{i}Y_{i}\right|\right]\le\sum_{i=1}^{n}\sqrt{\mathbb{E}[X_{i}^{2}]}\sqrt{\mathbb{E}[Y_{i}^{2}]}\le\sqrt{\sum_{i=1}^{n}\mathbb{E}[X_{i}^{2}]\sum_{i=1}^{n}\mathbb{E}[Y_{i}^{2}]}\to0,
\]
and thus Markov's inequality implies $\sum_{i=1}^{n}X_{i}Y_{i}\stackrel{p}{\to}0$.
Combining these results yields the conclusion. \qed \\

The following lemma restates the general CLT for infinite-order generalized $U$-statistics in \citet*[Theorem 2]{peng2022rates}. For other CLT results for infinite-order $U$-statistics under various different conditions, see, \cite{mentch2016quantifying}, \cite{wager2018estimation}, and \cite{diciccio2022clt}.
\begin{lemma}[CLT for incomplete generalized $U$-statistics]  \label{lem:CLT}
Let $(Z_1,\dots,Z_n)$ be i.i.d. random variables, $(\rho_\iota)_{\iota\in I_n}$ i.i.d. Bernoulli random variables with $N_{n}=\mathbb{E}[B_{n}]=p_{n}|I_{n}|$, $p_{n}=\mathbb{P}\{\rho_{\iota}=1\}$ and independent from $(\rho_\iota)_{\iota\in I_n}$, and
\[
\hat \theta_n = \frac{1}{B_n}\sum_{\iota \in I_{n}} \rho_{\iota}h_n(Z_\iota,\omega_\iota)
\]
be an incomplete generalized $U$-statistic with a kernel $h_n: \mathcal{Z}^{s_n} \times \Omega \to \mathbb{R}$ and $B_n=\sum_{\iota \in I_n}\rho_\iota$. Define $\theta_n = \mathbb{E}[h_n(Z_\iota,\omega_\iota)]$, $\zeta_{n,1}=\Var(\mathbb{E}[h_n(Z_\iota,\omega_\iota)| Z_1])$, and $\zeta_{n,s} =\Var(h_n(Z_\iota,\omega_\iota))$. Suppose that, as \(n\to\infty\),
\[
\frac{\mathbb{E}[| h_n(Z_\iota,\omega_\iota) - \theta_n|^{2k}]}{\{\mathbb{E}[| h_n(Z_\iota,\omega_\iota) - \theta_n|^k]\}^2}=O(1),
\]
for $k=2,3$,  \(N_n\to\infty\), and the variance-ratio satisfies \[ \frac{s}{n}\frac{\zeta_{n,s}}{s\zeta_{n,1}} =o(1). \]
Then we have
\[
\frac{\hat{\theta}_n - \theta_n}{\sqrt{s^2\zeta_{n,1}/n +\zeta_{n,s}/N_n}} \xrightarrow{d} \mathcal{N}(0,1).
\]
\end{lemma}


\acks{%
This research was supported in part by JSPS KAKENHI grant 23K01331.
All remaining errors are ours.
}
\bibliography{bib}

\end{document}